%% file: final.tex
\documentclass[acmsmall]{acmart}
\usepackage{multirow}
\usepackage{colortbl}
\usepackage{makecell}
\usepackage{xcolor}
\usepackage{xspace}
\usepackage{subcaption}
\usepackage{enumitem}
\usepackage{tcolorbox}
\usepackage{cleveref}
\usepackage{pifont}
\tcbuselibrary{skins,breakable}
\usepackage{wrapfig}

\definecolor{orange-web}{RGB}{255, 165, 0}      
\definecolor{sagegreen}{RGB}{138, 179, 137}     
\definecolor{lemonyellow}{RGB}{255, 247, 0}     
\definecolor{skyblue}{RGB}{135, 206, 235}       
\definecolor{coral}{RGB}{255, 127, 80}          
\definecolor{lavender}{RGB}{230, 230, 250}      
\definecolor{mintgreen}{RGB}{152, 255, 152}     
\definecolor{peach}{RGB}{255, 218, 185}         
\definecolor{steelblue}{RGB}{70, 130, 180}      
\definecolor{rosegold}{RGB}{183, 110, 121}      

\colorlet{boxcolor}{sagegreen}  

\newtcolorbox{findingbox}[1][]{
  enhanced,
  attach boxed title to top left={xshift=4mm,yshift=-2mm},
  colback=boxcolor!10,
  colframe=boxcolor!60,
  colbacktitle=boxcolor!80,
  coltitle=white,
  fonttitle=\bfseries\small,
  boxed title style={size=small, colframe=boxcolor!80, sharp corners},
  sharp corners,
  boxrule=0.8pt,
  left=4pt, right=4pt, top=4pt, bottom=4pt,
  breakable,
  title={#1}
}

\setcopyright{cc}
\setcctype{by}
\acmDOI{10.1145/3832167}
\acmYear{2026}
\acmJournal{PACMSE}
\acmVolume{3}
\acmNumber{ISSTA}
\acmArticle{ISSTA076}
\acmMonth{10}
\acmSubmissionID{issta26main-p726-p}
\received{2026-01-30}
\received[accepted]{2026-04-16}

\newcommand{\nbc}[3]{
	{\colorbox{#3}{\bfseries\sffamily\scriptsize\textcolor{white}{#1}}}
	{\color{#3}{\small{#2}}}}

\newcommand{\option}[1]{\nbc{OPTION #1}{}{violet!70!black}}

\definecolor{optionbg1}{RGB}{255, 245, 238}  
\definecolor{optionbg2}{RGB}{240, 255, 240}  
\definecolor{optionbg3}{RGB}{240, 248, 255}  

\newtcolorbox{optionbox}[1][1]{
  enhanced,
  colback=optionbg#1,
  colframe=violet!40,
  boxrule=0.5pt,
  left=4pt, right=4pt, top=2pt, bottom=2pt,
  breakable,
  before upper={\option{#1}\hspace{2mm}}
}

\newcommand{\ourmethod}{\textsc{CodeOCR}\xspace}

\begin{document}

\title{Seeing Is Coding: On the Effectiveness of Vision Language Models in Code Understanding}

\author{Yuling Shi}
\orcid{0009-0009-9738-7072}
\affiliation{%
  \institution{Shanghai Jiao Tong University}
  \city{Shanghai}
  \country{China}
}
\email{yuling.shi@sjtu.edu.cn}
\author{Chaoxiang Xie}
\orcid{0009-0001-7168-6332}
\affiliation{%
  \institution{Hohai University}
  \city{Nanjing}
  \country{China}
}
\email{edxxx1@icloud.com}
\author{Zhensu Sun}
\orcid{0000-0001-5393-7858}
\affiliation{%
  \institution{Singapore Management University}
  \city{Singapore}
  \country{Singapore}
}
\email{zssun@smu.edu.sg}
\author{Yeheng Chen}
\orcid{0009-0001-6800-0344}
\affiliation{%
  \institution{Shanghai Jiao Tong University}
  \city{Shanghai}
  \country{China}
}
\email{yeheng1@ualberta.ca}
\author{Chenxu Zhang}
\orcid{0009-0007-6516-4945}
\affiliation{%
  \institution{Imperial College London}
  \city{London}
  \country{United Kingdom}
}
\email{cz3625@ic.ac.uk}
\author{Longfei Yun}
\orcid{0009-0002-3652-4095}
\affiliation{%
  \institution{University of California at San Diego}
  \city{San Diego}
  \country{USA}
}
\email{loyun@ucsd.edu}
\author{Chengcheng Wan}
\orcid{0000-0001-9162-9688}
\affiliation{%
  \institution{East China Normal University}
  \city{Shanghai}
  \country{China}
}
\affiliation{%
  \institution{Shanghai Innovation Institute}
  \city{Shanghai}
  \country{China}
}
\email{ccwan@sei.ecnu.edu.cn}
\author{Hongyu Zhang}
\orcid{0000-0002-3063-9425}
\affiliation{%
  \institution{Chongqing University}
  \city{Chongqing}
  \country{China}
}
\email{hyzhang@cqu.edu.cn}
\author{David Lo}
\orcid{0000-0002-4367-7201}
\affiliation{%
  \institution{Singapore Management University}
  \city{Singapore}
  \country{Singapore}
}
\email{davidlo@smu.edu.sg}
\author{Xiaodong Gu}
\orcid{0000-0002-0529-6408}
\affiliation{%
  \institution{Shanghai Jiao Tong University}
  \city{Shanghai}
  \country{China}
}
\email{xiaodong.gu@sjtu.edu.cn}
\renewcommand{\shortauthors}{Shi et al.}

\begin{abstract}

Large Language Models (LLMs) have achieved remarkable success in source code understanding, yet as software systems grow in scale, computational efficiency has become a critical bottleneck.
Currently, these models rely on a text-based paradigm that treats source code as a linear sequence of tokens, which leads to a linear increase in context length and associated computational costs.
The rapid advancement of Multimodal LLMs (MLLMs) introduces an opportunity to optimize efficiency by representing source code as rendered images.
Unlike text, which is difficult to compress without losing semantic meaning, the image modality is inherently suitable for compression.
By adjusting resolution, images can be scaled to a fraction of their original token cost while remaining recognizable to vision-capable models.
To explore the feasibility of this approach, we conduct the first systematic study on the effectiveness of MLLMs for code understanding.
Our experiments reveal that: (1) MLLMs can effectively understand code with substantial token reduction, achieving up to 8× compression;
(2) MLLMs can effectively leverage visual cues such as syntax highlighting, improving code completion performance under 4× compression; and (3) Code-understanding tasks like clone detection exhibit exceptional resilience to visual compression, with some compression ratios even slightly outperforming raw text inputs.
Our findings highlight both the potential and current limitations of MLLMs in code understanding, which points out a shift toward image-modality code representation as a pathway to more efficient inference.
\end{abstract}

\ccsdesc[500]{Software and its engineering~Software prototyping}
\ccsdesc[500]{Software and its engineering}

\keywords{Vision Language Models, Code Understanding, Multimodal Learning, Software Engineering}

\maketitle

\section{Introduction}
LLMs have established a dominant paradigm in software engineering~\cite{chen2021codex,roziere2023codellama,fan2023large,shi2024between,jiang2024survey}.
Currently, these models primarily operate on a text-based paradigm, where source code is treated as a linear sequence of tokens~\cite{zhang2024unifying}.
However, as software systems grow in scale and complexity, the resulting linear increase in context length and its associated computational overhead has become a significant efficiency bottleneck~\cite{guo2023longcoder,bogomolovLongCodeArena2024,shi2026reasoning,wang2026fasa,wang2025position}.

\begin{figure}[ht]
\centering
\includegraphics[width=0.95\linewidth]{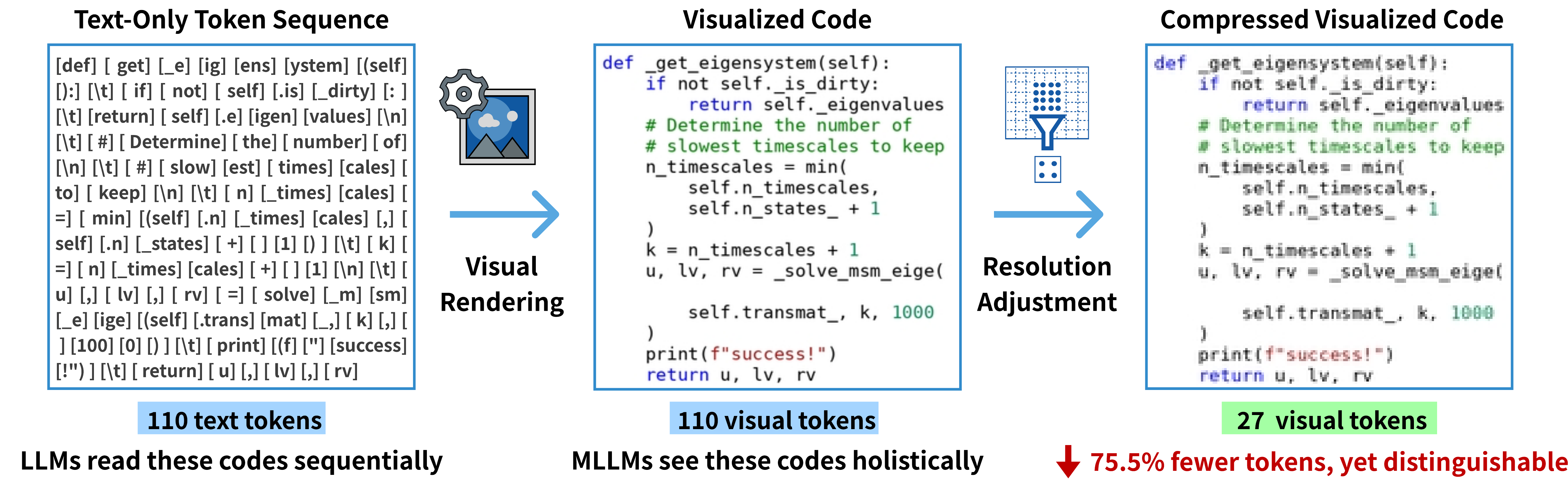}
\vspace{-0.25cm}
\caption{An Example of Code Representations across Different Modalities.}
\vspace{-0.25cm}
\label{fig:motivation}
\end{figure}

The rapid advancement of multimodal LLMs---particularly Vision Language Models (VLMs) that integrate visual understanding capabilities~\cite{openai2023gpt4,team2023gemini,liu2024llava,yin2024survey,ma2026llm,guo2026shredbench}---presents a promising opportunity to mitigate this limitation.
To be specific, many popular LLMs, such as GPT-5~\cite{openai_gpt5mini} and Gemini-3~\cite{gemini3flash_model_card_2025}, now natively support multimodal inputs, enabling the processing of text and visual data within a unified architecture.
This capability motivates us to reconsider how source code can be represented.
Compared with text, image modality exhibits a key advantage in terms of compressibility~\cite{li2025text}.
Image data can be scaled by simply adjusting resolution~\cite{wei2025deepseekocr,li2025text}, whereas compressing code text for LLMs is a discrete and often lossy process involving token pruning or semantic rewriting.
As a result, representing source code as rendered images (i.e., code images) could provide a more scalable and computationally efficient alternative to traditional text representations.
Consider the code snippet in \Cref{fig:motivation} for example.
This snippet, when represented as text (left), costs approximately 110 text tokens.
When rendered into a code image (middle), its resolution is calibrated to occupy an equivalent budget of 110 visual tokens (image tokens are priced at standard text token rates for vision-capable models like GPT-5 and Gemini-3 ~\cite{openai_api_pricing_2025,gemini_api_pricing_2025}).
However, this code image can be further compressed by reducing its resolution, yielding a 75.5\% reduction to just 27 visual tokens while maintaining clarity in readability. 
In contrast, text-based compression methods that achieve similar reduction ratios typically rely on token pruning, which often result in significant information loss~\citep{shi2025longcodezip,zhang2022dietcode,pan2025hiddencost}.
This flexibility highlights the potential of code images to alleviate the high inference costs and context-window constraints faced by current LLMs~\cite{jiang2023llmlingua,shi2025longcodezip}.

To leverage this advantage, a fundamental question naturally arises: how effectively can LLMs understand and reason over code images?
The answer to this question may signal a paradigm shift in how source code should be represented for AI to understand.
However, to the best of our knowledge, this question remains largely unexplored within the research community.
Existing research~\cite{you2024ferretui,baechler2024screenai,cheng2024guiworld,yang2025ui2code} has largely focused on GUI understanding and UI-to-code generation, where the visual inputs are graphical user interfaces rather than code images.
While these works demonstrate that LLMs can exhibit strong coding capabilities when processing visual inputs, they do not investigate whether code images themselves constitute a viable or effective representation of source code.
To address this question, a large-scale, comprehensive experimental evaluation of LLMs' performance on code images is necessary.

To fill this knowledge gap, we conduct a comprehensive empirical study on image-based code representation with LLMs guided by five research questions.
In the study, seven widely used LLMs, all of which support multimodal inputs, are evaluated across 4 downstream tasks (code completion, code summarization, clone detection, and code question answering), while systematically investigating compression ratios (1×--8×) and rendering strategies (plain, highlighted, bolded).
Next, we introduce the five RQs and summarize essential findings.

\smallskip

\noindent \textbf{RQ1: How effective are LLMs in visual code understanding compared with textual code?}
As the first step, we investigate whether LLMs can process visualized code as effectively as traditional text.
Specifically, for each sample in all benchmarks, we derive a variant by rendering it into images that contain the same number of tokens. 
We then compare model performance when the input is provided as raw text versus as a code image

\noindent \textit{Findings.}
For all the four downstream tasks, LLMs with visualized code input can achieve comparable or even superior performance to textual input. For example, GPT-5-mini achieves 42\% F1 improvement with code images over raw text for clone detection, and Gemini-3-Pro demonstrates comparable or superior performance across all four tasks.
These results indicate that replacing textual code with visual representations is both viable and promising for code understanding, demonstrating the feasibility of leveraging the multimodal capabilities of modern LLMs.
However, performance improvements are not uniform across models and tasks, suggesting that current LLMs are not yet fully optimized for this paradigm.
Bridging this gap remains an important direction for future research.


\smallskip

\noindent \textbf{RQ2: How resilient are LLMs to visual compression across different coding tasks?}
Building on the feasibility established in RQ1, we further explore a key advantage of visual representations, i.e., their compressibility.
Specifically, we vary the compression ratios of code images from 1× to 8× and systematically evaluate LLM performance under each compression ratio.


\noindent \textit{Findings.}
LLMs can exhibit exceptional compression resilience across tasks, with multiple models exceeding raw text baseline even at 8× compression ratio, i.e., costing only 12.5\% of tokens for raw text input.
For example, Gemini-3-Pro achieves 79.5\% accuracy on code question answering at 8× compression, surpassing its 74.8\% raw text baseline.
These results effectively demonstrate a remarkable robustness of LLMs' understanding of code images, highlighting a key advantage of image-based representations over linear text.
Consistent with RQ1, compression resilience varies across tasks and models, with state-of-the-art models like Gemini-3 and GPT-5 series maintaining stable or improved performance under nearly all compression settings.
This suggests that robust visual compression understanding is an achievable capability, pointing to clear opportunities for future model development.


\smallskip

\noindent \textbf{RQ3: Can visual enhancements (e.g., syntax highlighting, bold rendering) further improve LLMs' understanding of code images?}
In the previous RQs, we established that code images are a viable and compressible medium.
However, another key advantage of the visual modality is the ability to incorporate visual cues that are absent in raw text.
In this RQ, we investigate whether visual enhancements, specifically syntax highlighting and bolding, provide benefits to LLMs.

\noindent \textit{Findings.}
Visual enhancements improve model performance primarily at moderate compression levels (1×–4×), with diminishing returns at higher ratios.
Both syntax highlighting and bold rendering provide consistent gains when the underlying visual signal remains legible---at 1×–2× compression, multiple models show 1–3\% improvements in Edit Similarity and accuracy.
However, at 8× compression, these enhancements offer limited benefit, as reduced resolution obscures the visual distinctions they introduce.
Notably, bold rendering can exacerbate degradation at extreme compression ratios by further reducing character clarity.
These findings indicate that visual enhancements are most effective within a compression ``sweet spot,'' motivating future work on adaptive rendering strategies.


\smallskip

\noindent \textbf{RQ4: Can LLM's understanding of visualized code generalize across programming languages?}
To ensure our findings are not Python-specific, we replicate the experiments for RQ1--RQ3 on Java.

\noindent \textit{Findings.}
The core trends remain consistent across languages.
The Gemini family achieves up to 12\% ES improvement in Java code completion, and clone detection shows 6--20\% ACC gains with visual inputs across multiple models.
Model-specific strengths and compression resilience patterns also hold: models that performed well under compression in Python maintain their relative advantages in Java.
These results support that the core findings generalize across programming languages.

\smallskip

\noindent \textbf{RQ5: How does visual compression degrade code information, and what error types emerge across compression ratios?}
To better understand how information is lost under visual compression, we conduct a detailed degradation analysis.
Specifically, we perform OCR-style code reconstruction experiments, in which LLMs are required to reproduce the code content from compressed code images across compression ratios ranging from 1× to 8×.
We then analyze the errors between the original and reconstructed code.

\noindent \textit{Findings.}
Information degradation follows a clear hierarchical pattern.
Token-level errors emerge first at low compression (1×–2×), followed by line-level errors at moderate compression (2×–4×), while block-level errors dominate under high compression (4×–8×).
The 4×--8× range represents a critical threshold---most models experience significant degradation, while the Gemini-3 family maintains stability with high CodeBLEU even at 8× compression. Crucially, we found that the token-level errors do not always impair downstream semantic performance, suggesting that LLMs can often infer the correct logic even when the visual signal is slightly blurred.

\smallskip

\begin{figure}[t]
\centering
\includegraphics[width=0.95\linewidth]{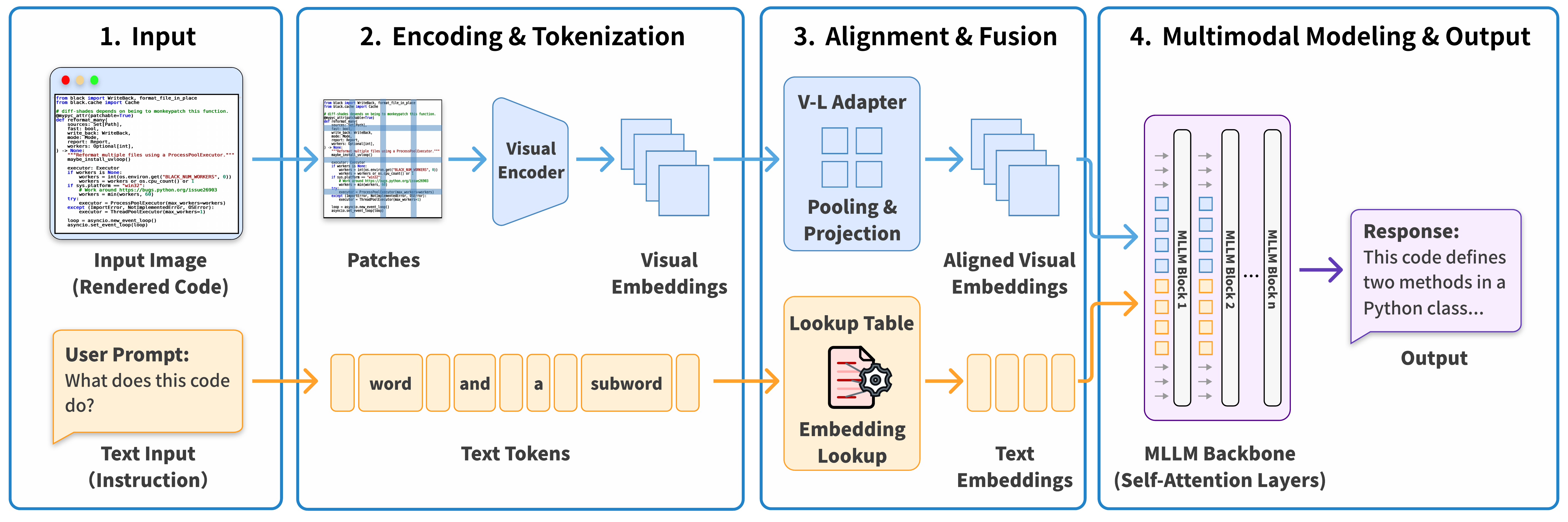}
\vspace{-0.25cm}
\caption{Multimodal Processing Pipeline for Visualized Code Understanding in MLLMs.}
\vspace{-0.5cm}
\label{fig:mllm}
\end{figure}

These empirical results provide valuable insights for developing code image understanding systems.
To this end, we implement \ourmethod~\cite{codeocr_repo}, a practical tool for rendering source code into images with configurable visual enhancements and compression ratios for researchers or developers to use.
To sum up, this paper makes the following contributions:
\begin{itemize}[leftmargin=0.5cm]
    \item We perform the first comprehensive empirical study on visual code understanding, evaluating seven state-of-the-art MLLMs across four downstream tasks with systematic analysis of compression ratios and rendering strategies.
    \item We empirically demonstrate that image-based code representation is a viable technical direction, where, without any targeted optimization, multiple existing LLMs can achieve comparable or even superior performance to text-based baselines on code understanding tasks.
    \item We propose and implement a practical tool called \ourmethod for rendering source code into image, supporting LLMs to process code in a more token-efficient manner.
\end{itemize}

\section{Background}
\label{sec:background}

Multimodal capability has become a native feature in state-of-the-art LLMs like GPT-5 and Gemini-3, enabling them to process both text and images within a unified architecture~\cite{openai_gpt5mini,gemini3pro_model_card_2025,bai2025qwen3vltechnicalreport,vteam2026glm45v,chen2025autoneural}.  

But how do these models process visual inputs? We illustrate the pipeline in Figure~\ref{fig:mllm}, which consists of four stages.

\textbf{Stage 1: Inputs.} The code is rendered as an image $I \in \mathbb{R}^{H \times W \times 3}$ with visual cues like syntax highlighting and indentation (left panel). Alongside, a text prompt provides the instruction. While text-based models directly tokenize raw code strings, MLLMs treat code as a 2D visual artifact.

\textbf{Stage 2: Encoding \& Tokenization.} The rendered image is divided into fixed-size patches (e.g., 14×14 pixels). A visual encoder (typically a Vision Transformer) converts these patches into visual embeddings:
\begin{equation}
V = \text{Encoder}(I) = \{v_1, v_2, \dots, v_N\}
\end{equation}
where each $v_i$ captures the visual features of a patch. In parallel, the text prompt is tokenized into a sequence of text tokens (words or subwords).

\textbf{Stage 3: Alignment \& Fusion.} The visual and text tokens are processed through separate alignment modules before fusion. For visual tokens, a V-L Adapter applies pooling and projection to compress adjacent patches into aligned visual embeddings. For instance, a $2 \times 2$ pooling operation merges four patches:
\begin{equation}
T_v = \text{MLP}(\text{Concat}(v_{i,j}, v_{i+1,j}, v_{i,j+1}, v_{i+1,j+1}))
\end{equation}
This reduces the number of visual tokens while preserving semantic density. For text tokens, a lookup table maps each token to its corresponding text embedding. The aligned visual embeddings and text embeddings are then concatenated to form a unified input sequence.

\begin{figure}[t]
\centering
\includegraphics[width=0.95\linewidth]{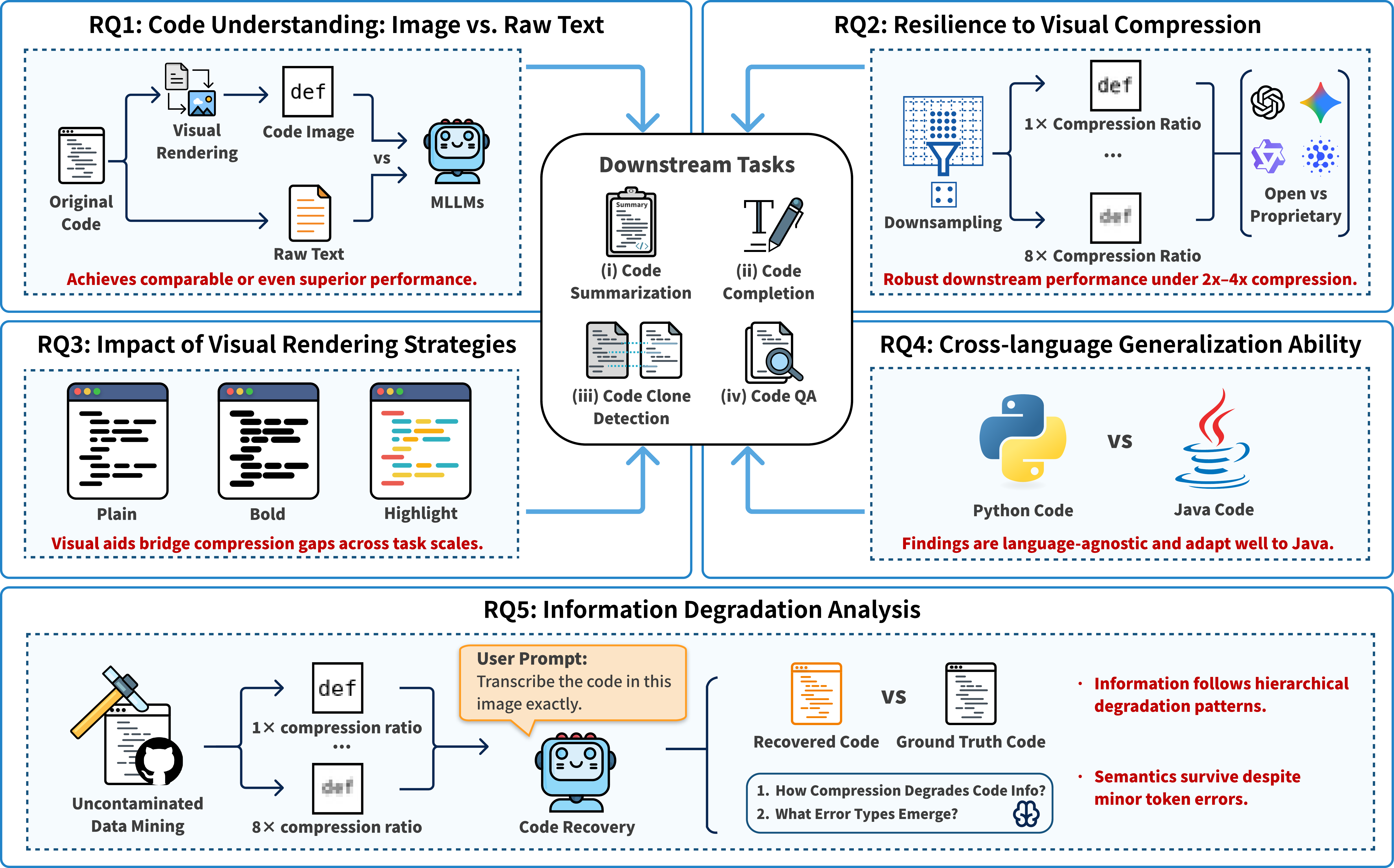}
\caption{Overview of the Empirical Study Design and Core Findings.}
\label{fig:overview}
\vspace{-0.2cm}
\end{figure}

\textbf{Stage 4: Multimodal Modeling \& Output.} The MLLM backbone (self-attention layers) processes the unified sequence:
\begin{equation}
\text{Input} = [T_v; T_{text}]
\end{equation}
Unlike text models that rely on discrete vocabulary and syntax rules, MLLMs learn to interpret continuous visual patterns—such as color-coded keywords, indentation depth, and bracket alignment—directly from pixel data. This enables them to understand code structure without explicit parsing, leveraging the same spatial reasoning used for natural images. 

Importantly, this multimodal capability is additive rather than a trade-off—multimodal variants maintain comparable performance to text-only counterparts (e.g. Qwen-3-VL vs Qwen-3) on NLP and coding benchmarks~\cite{bai2025qwen3vltechnicalreport,vteam2026glm45v}. This visual processing capability opens new possibilities for representing structured content—including source code—as images rather than text tokens, which we systematically investigate in this paper.


\section{Experimental Setting}
In this section, we introduce the experimental setting for this study, including the task, models, benchmark, evaluation metrics, and implementation details.
Our study is guided by five research questions:
\begin{itemize}[leftmargin=*]
    \item \textbf{RQ1:} How effective are LLMs in visual code understanding compared with textual code?
    \item \textbf{RQ2:} How resilient are LLMs to visual compression across different coding tasks?
    \item \textbf{RQ3:} Can visual enhancements (e.g., syntax highlighting, bold rendering) further improve LLMs' understanding of code images?
    \item \textbf{RQ4:} Can LLM's ability of visual code understanding generalize across programming languages?
    \item \textbf{RQ5:} How does visual compression degrade code information, and what error types emerge across compression ratios?
\end{itemize}

An overview of our experimental design is presented in Figure~\ref{fig:overview}. Our investigation follows a progressive logic: we begin by establishing the fundamental feasibility of visual code understanding compared to text (RQ1). Building on this baseline, we explore the core advantage of the visual modality—optical compression—by systematically varying resolution (RQ2). To further optimize performance, we examine whether visual cues like syntax highlighting can mitigate compression loss (RQ3). Finally, we validate the robustness of our findings across different programming languages (RQ4) and conduct a microscopic analysis of information degradation patterns (RQ5).



\subsection{Benchmark and Metrics}\label{sec:bm}

\begin{table}[t]
\centering
\caption{Summary of Tasks Used in Our Evaluation.}
\vspace{-0.25cm}
\label{tab:datasets}
\resizebox{0.65\linewidth}{!}{
\begin{tabular}{llccc}
\toprule
\textbf{Task} & \textbf{Language} & \textbf{\# Examples} & \textbf{Avg. Context Len.} & \textbf{Avg. GT Len.} \\
\midrule
Code Summarization               & Python & 109 & 6184.1 & 1481.8 \\
\midrule
\multirow{2}{*}{Code Completion} & Python & 200 & 6138.6 & 12.3 \\
                                 & Java   & 200 & 5653.5 & 11.9  \\
\midrule
\multirow{2}{*}{Code Clone Detection} & Python & 200 & 124.9 & 1.0 \\
                                      & Java   & 200 & 215.7 & 1.0 \\
\midrule
Code Question Answering               & Python & 200 & 1316.9 & 1.0 \\
\bottomrule
\end{tabular}
}
\vspace{-0.5cm}
\end{table}


To comprehensively evaluate visual code understanding, we select four representative tasks spanning different levels of code comprehension.
All tasks require models to process code as input, aligning with our focus on code understanding capability.
We primarily evaluate on Python and extend our analysis to Java in RQ4.
Dataset statistics are summarized in Table~\ref{tab:datasets} and the lengths are computed with the tokenizer of Qwen-3-VL~\citep{bai2025qwen3vltechnicalreport}.
\textbf{Code Completion} tests fine-grained syntactic understanding.
We adopt the LongCodeCompletion dataset~\citep{guo2023longcoder} and randomly sample 200 Python and 200 Java samples from the challenging subset curated by~\citet{shi2025longcodezip}.
We apply Retrieval-Augmented Generation (RAG) to provide relevant code context (details in Section~\ref{sec:implementation}).
The average context lengths are 6,139 tokens for Python and 5,654 tokens for Java.
We use \textbf{Exact Match (EM)} and \textbf{Edit Similarity (ES)}~\citep{guo2023longcoder} for evaluation: EM measures whether the generated code exactly matches the ground truth, while ES captures partial correctness via token-level Levenshtein distance.

\textbf{Code Summarization} evaluates high-level semantic extraction.
We use the LongModuleSummarization dataset~\citep{bogomolovLongCodeArena2024} following~\citep{shi2025longcodezip,zeng2026readability}, containing 109 examples with an average of 6,184 tokens per sample.
We adopt \textbf{CompScore}~\citep{bogomolovLongCodeArena2024}, an LLM-as-judge metric where DeepSeek-V3.2~\citep{deepseekai2024deepseekv3} compares generated documentation against ground truth with bidirectional averaging to mitigate ordering bias (scores range 0--100, where 50 indicates parity).

\textbf{Code Clone Detection} assesses semantic similarity recognition.
We employ GPTCloneBench~\citep{gptclonebench2023}, focusing on Type-4 (semantic) clones---code pairs implementing identical functionality with different syntax and structure.
For each language (Python and Java), we randomly sample a balanced dataset of 200 pairs (100 positive, 100 negative).
The average context lengths are 125 tokens for Python and 216 tokens for Java.
We report \textbf{Accuracy (ACC)} and \textbf{F1 score}, where F1 provides a more balanced assessment given the class imbalance.

\textbf{Code Question Answering} examines code comprehension through question answering, where models must select the correct answer from multiple choices based on the provided code context.
We construct a dataset following the format of LongCodeQA~\citep{rando2025longcodebench}, as our preliminary experiments revealed severe data leakage in the original dataset---GPT-5.1 achieved 81.5\% accuracy when given only questions and options without any code context, indicating that models could answer correctly through memorization rather than code understanding.
To address this, we crawled 35 Python repositories from GitHub (created after August 2025, 10+ stars) and used DeepSeek V3.2~\citep{deepseekai2024deepseekv3} to generate an initial pool of 1,000 candidate (context, question) pairs following the original dataset's exact format.
Three PhD students unaffiliated with the authors, each with 3+ years of programming experience, then validated each question one by one, ensuring: (1) the question is meaningful and valuable for evaluating code comprehension, (2) the question is answerable from the provided code context, (3) the context is necessary to determine the correct answer, and (4) exactly one answer is unambiguously correct.
Only questions receiving unanimous approval from all three validators were retained, and annotation continued until 200 validated samples were collected.
Finally, the authors shuffled answer option orders to avoid positional bias.
This curated dataset is publicly available for research use.
We evaluate performance using \textbf{Accuracy (ACC)}.

\begin{table}[t]
\centering
\caption{Summary of Evaluated MLLMs with Release Information and API Pricing (per 1M tokens).}
\vspace{-0.3cm}
\label{tab:models}
\resizebox{0.85\linewidth}{!}{
\begin{tabular}{l|c|c|c|c|cc|cc}
\toprule
\multirow{2}{*}{\textbf{Model}} & \textbf{Release} & \textbf{Knowledge} & \multirow{2}{*}{\textbf{Type}} & \textbf{Multimodal} & \multicolumn{2}{c|}{\textbf{Price ($\le$ 200k)}} & \multicolumn{2}{c}{\textbf{Price ($>$ 200k)}} \\
\cline{6-9}
 & \textbf{Date} & \textbf{Cut-off} & & \textbf{Ability} & \textbf{Input} & \textbf{Output} & \textbf{Input} & \textbf{Output} \\
\midrule
Qwen-3-VL & 2025-09 & 2024-06 & Open-weight & \textcolor{green!70!black}{\ding{51}} & \$0.40 & \$1.60 & \$0.40 & \$1.60 \\
GLM-4.6v & 2025-12 & 2025-07 & Open-weight & \textcolor{green!70!black}{\ding{51}} & \$0.30 & \$0.90 & \$0.30 & \$0.90 \\
GPT-5-mini & 2025-08 & 2024-05 & Proprietary & \textcolor{green!70!black}{\ding{51}} & \$0.25 & \$2.00 & \$0.25 & \$2.00 \\
GPT-5.1 & 2025-11 & 2024-09 & Proprietary & \textcolor{green!70!black}{\ding{51}} & \$1.25 & \$10.00 & \$1.25 & \$10.00 \\
Gemini-2.5-Pro & 2025-06 & 2025-01 & Proprietary & \textcolor{green!70!black}{\ding{51}} & \$1.25 & \$10.00 & \$2.50 & \$15.00 \\
Gemini-3-Flash & 2025-12 & 2025-01 & Proprietary & \textcolor{green!70!black}{\ding{51}} & \$0.50 & \$3.00 & \$0.50 & \$3.00 \\
Gemini-3-Pro & 2025-11 & 2025-01 & Proprietary & \textcolor{green!70!black}{\ding{51}} & \$2.00 & \$12.00 & \$4.00 & \$18.00 \\
\bottomrule
\end{tabular}
}
\vspace{-0.4cm}
\end{table}

\subsection{Studied Large Language Models}\label{sec:llms}
To ensure the generalizability of our findings, we evaluate seven state-of-the-art LLMs with multimodal capability spanning both proprietary and open-weight categories.
Table~\ref{tab:models} summarizes model details and official pricing as of January 30, 2026~\citep{openrouter}.
The proprietary models include GPT-5-mini and GPT-5.1~\citep{openai_gpt5mini,openai_gpt51} from OpenAI, and Gemini-2.5-Pro, Gemini-3-Flash, and Gemini-3-Pro~\citep{comanici2025gemini25pushingfrontier,gemini3flash_model_card_2025,gemini3pro_model_card_2025} from Google.
For open-weight models, we include Qwen-3-VL with 235B parameters~\citep{bai2025qwen3vltechnicalreport} and GLM-4.6v with 108B parameters~\citep{vteam2026glm45v}, enabling reproducible research and architectural analysis.
Importantly, these proprietary models have multimodal capability natively integrated, while open-weight models have been officially benchmarked to match their text-only counterparts~\citep{bai2025qwen3vltechnicalreport,vteam2026glm45v}.
It ensures that our experimental setup does not introduce confounding factors from degraded baseline capability. 


\subsection{Visual Rendering of Source Code}\label{sec:visual_processing}

\begin{figure}[t]
    \centering
    \includegraphics[width=0.8\linewidth]{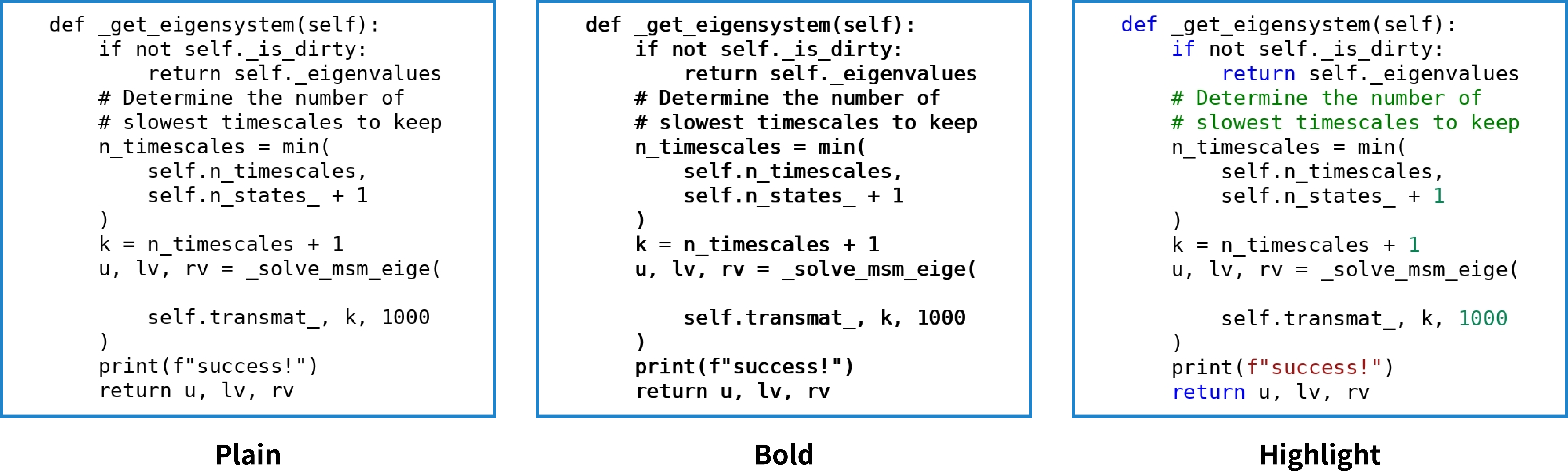}
    \vspace{-0.25cm}
    \caption{Examples of Visual Rendering Strategies: Plain, Bold, and Highlight.}
    \vspace{-0.5cm}
    \label{fig:RQ3Examples}
\end{figure}

\textbf{Code Rendering.}
We render source code into images at a high base resolution of 2240×2240 pixels, following prior work~\cite{liang2026visual}.
This resolution is selected for compatibility with modern MLLMs, as it is divisible by common image patch sizes (e.g., 14 and 16 pixels) used in visual encoders~\cite{bai2025qwen3vltechnicalreport}, ensuring that no partial patches are created during tokenization.
By default, we use plain rendering---black monospace text on a white background---which serves as the baseline configuration throughout our experiments.
To investigate the effect of visual enhancements (RQ3), we additionally support two variants: bold rendering with increased stroke width, and syntax highlighting following Visual Studio Code's~\cite{vscode} ``Default Light'' theme (Figure~\ref{fig:RQ3Examples}).
When the code exceeds a single page, we split it into multiple consecutive images while preserving line boundaries. Modern MLLMs natively support multi-image inputs and can process them in the provided order~\cite{bai2025qwen3vltechnicalreport, vteam2026glm45v}.

\noindent\textbf{Resolution Compression.}
MLLMs process images by dividing them into fixed-size patches and encoding each patch as visual tokens (Section~\ref{sec:background}).
For an image of resolution $W \times H$ with patch size $p$, the visual token count is $(W/p) \times (H/p)$.
We define compression ratio $k$× such that the visual token count equals exactly $1/k$ of the original text token count; thus, at 1× compression, the visual token count matches the text token count. And since providers typically price visual and text tokens at similar rates, this also results in comparable cost~\citep{openrouter}.
To generate images at any compression level, we start from the code image at the high base resolution (2240×2240), which produces more visual tokens than the equivalent text tokens, ensuring sufficient visual fidelity as the starting point.
We then apply bilinear downsampling to reach the exact target resolution corresponding to the desired $k$× compression.
In our experiments, we evaluate compression ratios of 1×, 2×, 4×, and 8× to investigate the trade-off between visual fidelity and token efficiency.

\subsection{Baselines and Input Design}
\label{sec:baseline}

\textbf{Input Modality.}
Following the established paradigm in visual text understanding research~\citep{wei2025deepseekocr, liang2026visual, zhao2025vtcbench}, we decouple code content from task instructions: code is rendered as images while instructions are provided in text form. This design enables us to isolate and evaluate the visual code understanding capability of MLLMs.
Specifically, for code summarization, the source code to be summarized is rendered as an image, accompanied by a text instruction requesting documentation generation; for code completion, the RAG-retrieved relevant code from the codebase is rendered as images while the incomplete code prefix and completion instruction remain in text; for clone detection, the two code snippets to be compared are rendered as separate images, with a text instruction asking the model to classify whether they are clones; for question answering, the code context is rendered as images while the question itself and answer options are provided in text.


\noindent\textbf{Baseline.}
We establish two baselines: (1) \textit{NoCtx} (No Context), where code context is removed and only the task instruction is kept to measure the lower bound and detect potential data leakage; and (2) \textit{Text}, where code is provided as plain text tokens, representing the standard text-based approach.
The NoCtx baseline is not applicable for Code Summarization and Clone Detection, as these tasks require source code to be summarized or compared.
\subsection{Implementation Details}\label{sec:implementation}

We implement our experiments in Python using a custom rendering pipeline built on Pygments~\cite{pygments} for syntax tokenization and Pillow~\cite{pillow} for image generation and processing.
The base images are rendered with the default monospaced font from Visual Studio Code~\cite{vscode}, at a font size of 40 pixels as suggested by prior work~\cite{liang2026visual, zhao2025vtcbench}, line height of 1.0, and margin of 1\% of the page width.
For syntax highlighting, we adopt the ``Default Light'' theme from Visual Studio Code~\cite{vscode}.
For bold rendering, following the font synthesis definitions in W3C CSS standards~\cite{w3c_css_fonts} and the FreeType engine~\cite{freetype}, we render each glyph multiple times with +1 pixel horizontal and vertical offsets to simulate increased stroke width.
Compression is achieved through bilinear downsampling with \texttt{Pillow}~\cite{pillow}.
For task-specific input preparation, most tasks directly use the code context with corresponding instructions according to Section~\ref{sec:bm}.
And code completion employs function-level Retrieval-Augmented Generation (RAG)~\cite{shi2025longcodezip} using UniXcoder~\cite{guo2022unixcoder} to retrieve the top-5 most similar code snippets as context.
In our experiments, all models are accessed through OpenRouter~\cite{openrouter}, a unified API gateway that provides standardized access to multiple LLM providers.
During inference, we use the default sampling parameters provided by the API provider and repeat all experiments for 5 times to report the average performance along with standard deviation.

\section{Results and Analysis}
In this section, we report our experimental results and answer the five research questions.

\subsection{RQ1: How Effective are LLMs in Understanding Visualized Code vs. Textual Code?}

In this RQ, we systematically evaluate whether LLMs can effectively understand code through visual representations.
To investigate this, we evaluate all seven models on the four Python tasks described in Section~\ref{sec:bm}, comparing their performance between raw text input and code image input. We also include the No Context baseline (``NoCtx'') defined in Section~\ref{sec:baseline} to rule out the possibility that models answer correctly through memorization rather than genuine code understanding.
We use the Wilcoxon signed-rank test~\cite{wilcoxon1945individual} to assess statistical significance between Text and Image inputs. The null hypothesis is that they exhibit no significant difference.
The results are presented in Table~\ref{tab:rq1_overall}.
We highlight cells where visualized input achieves better performance than textual input.
\input{tables/rq1.tex}

\subsubsection{Feasibility of Code Image Understanding}

For all four downstream tasks, LLMs with code images as input can achieve comparable or even superior performance to raw text, indicating that replacing textual code with visual representations is both viable and promising.

In \textit{code summarization}, the Gemini family achieves slightly higher CompScore with code images (e.g., Gemini-3-Pro: 56.0 $\rightarrow$ 56.8), with no statistically significant difference from text---indicating that visual representations preserve high-level semantic information.
In \textit{code completion}, Gemini-3-Flash (55.1 $\rightarrow$ 57.1) and Gemini-3-Pro (55.8 $\rightarrow$ 57.7) achieve significantly higher ES with code images ($p<0.05$).
In \textit{clone detection}, GPT-5-mini and GPT-5.1 show significant improvements ($p<0.01$): F1 increases by 42\% (33.2 $\rightarrow$ 47.0) and 33\% (46.8 $\rightarrow$ 62.4), respectively.
In \textit{code question answering}, Gemini-3-Flash (73.4 $\rightarrow$ 74.8) and Gemini-3-Pro (74.8 $\rightarrow$ 77.2) achieve significant gains ($p<0.05$).

Notably, Gemini-3-Pro demonstrates comparable or superior performance across all four tasks, suggesting that state-of-the-art LLMs can effectively leverage image-based code representations.
One possible explanation is that visual representations enable models to perceive code structure holistically---capturing indentation patterns, block boundaries, and long-range dependencies in a single glance---rather than processing tokens sequentially~\cite{storey2006theories,busjahn2015eye}.

\begin{findingbox}[Finding \#1]
For all four tasks, LLMs with code images can achieve comparable or even superior performance to raw text (e.g., GPT-5-mini achieves 42\% F1 improvement in clone detection), demonstrating the feasibility and promise of image-based code representation.
\end{findingbox}

\subsubsection{Model-Specific Variation}

However, performance improvements are not uniform across models and tasks.
We observe that stronger models tend to achieve better code image understanding effectiveness.

The Gemini-3 family demonstrates the most consistent results across tasks.
GPT-5-mini and GPT-5.1 show strong performance in clone detection, where visual inputs provide substantial improvements over text baselines.
In contrast, models such as Qwen-3-VL and GLM-4.6v exhibit significant degradation ($p<0.01$): Qwen-3-VL's ES in code completion drops from 49.7 to 35.5, while GLM-4.6v's clone detection accuracy decreases from 81.6 to 69.6.
This variation reveals that visual code understanding is not yet uniformly developed across model families, with significant optimization potential remaining for open-weight models.

To assess that models are genuinely leveraging visual code information instead of memorizing its training data, we compare against No Context baselines.
For models showing improvement, image performance substantially exceeds the No Context baseline.
For example, GLM-4.6v achieves 72.6\% accuracy in code QA with images, far above its No Context baseline of 37.2\%.
This confirms that these models are extracting meaningful information from visual code representations.

We also observe task-specific patterns.
Clone detection shows the most pronounced visual advantage, with GPT-5-mini and GPT-5.1 achieving statistically significant improvements ($p<0.01$).
We attribute this to the pairwise comparison nature of the task: visual representations may help models focus on high-level semantic patterns rather than being distracted by syntactic differences in token sequences.
Code summarization results show no significant differences between modalities, confirming that visual representations preserve high-level semantic information.
Code completion and question answering show greater model-dependent variation, reflecting the different demands these tasks place on code image understanding.

These observations suggest that current LLMs are not yet fully optimized for code image understanding.
Bridging this gap remains an important direction for future research.

\begin{findingbox}[Finding \#2]
Code image understanding effectiveness varies by model, with state-of-the-art models (Gemini-3 family) showing consistent results across tasks.
This variation suggests that current LLMs are not yet fully optimized for this paradigm, and bridging this gap remains an important direction for future research.
\end{findingbox}

\subsection{RQ2: How Resilient are LLMs to Visual Compression Across Different Coding Tasks?}

Building on the feasibility established in RQ1, we further explore a key advantage of visual representations, i.e., their compressibility.
Specifically, we vary the compression ratios of code images from 1× to 8× and systematically evaluate LLM performance under each compression level.
We apply the Wilcoxon signed-rank test~\cite{wilcoxon1945individual} to assess whether performance under compression differs significantly from the uncompressed baseline (1×), with the null hypothesis that there is no significant difference.

\begin{figure}[t]
    \centering
    \includegraphics[width=\linewidth]{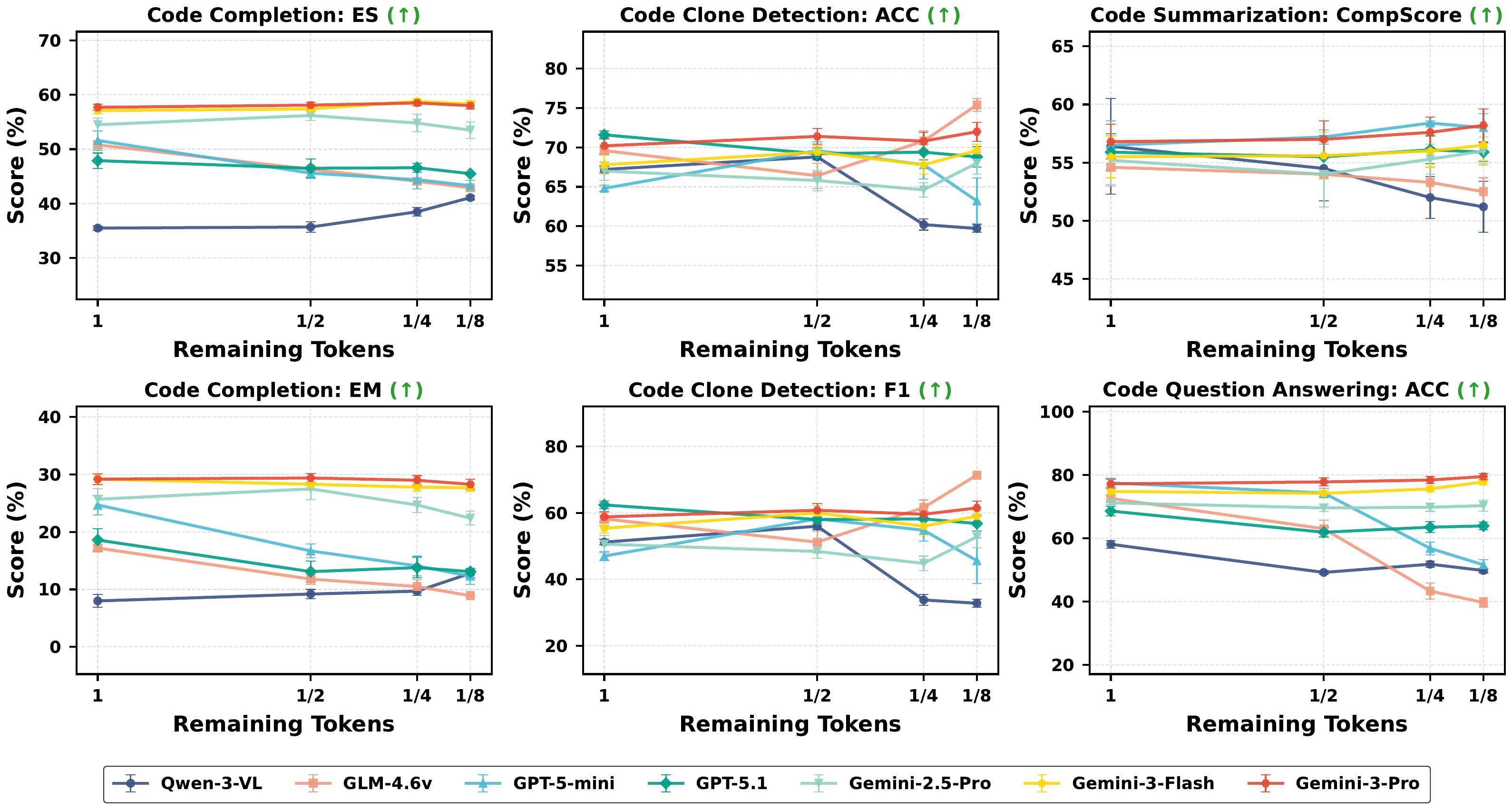}
    \vspace{-0.3cm}
    \caption{Performance under Varying Remaining Tokens across Different Tasks.} 
    \vspace{-0.3cm}
    \label{fig:RQ2}
\end{figure}

\subsubsection{Compression Effects Across Tasks}

We observe that compression resilience varies across tasks, with some tasks tolerating higher compression ratios than others.
In \textit{code summarization}, several models maintain or even improve performance under compression.
GPT-5-mini peaks at 4× compression with significant improvement (58.4 vs. 57.1 raw text, $p<0.05$), and Gemini-3-Pro improves from 56.0 (raw text) to 58.2 at 8×.
In contrast, weaker models such as Qwen-3-VL and GLM-4.6v show consistent degradation as compression increases.
In \textit{clone detection}, GPT-5-mini shows significant improvement with compressed images ($p<0.01$): F1 increases by 75\% from 33.2 (raw text) to 58.2 at 2× compression.
One possible explanation is that moderate compression acts as a denoising mechanism, blurring syntactic details and encouraging models to focus on semantic equivalence rather than surface-level differences.
In \textit{code completion}, performance varies substantially across models.
The Gemini-3 family significantly outperforms raw text across all compression levels ($p<0.05$; Gemini-3-Flash: ES 57.1--58.8 vs. 55.1 raw text), while other models show significant degradation.
Notably, Qwen-3-VL's ES increases from 35.5 at 1× to 41.1 at 8×.
We attribute this to the model's limited code image understanding capability: as compression reduces image clarity, the visual input provides less ``interference,'' and performance converges toward the no-context baseline (ES 45.0).
In \textit{code question answering}, the Gemini-3 family demonstrates significant improvements under compression ($p<0.05$; Gemini-3-Pro: 77.2 at 1× to 79.5 at 8× vs. 74.8 raw text), while other models exhibit significant degradation at higher compression ratios.
This resilience may stem from two factors: (1) modern MLLMs are trained on diverse image resolutions, developing inherent robustness to visual degradation~\cite{bai2025qwen3vltechnicalreport}; and (2) LLMs' strong language priors enable them to infer missing details from partial visual signals, similar to how humans read blurred text by leveraging contextual expectations.


\begin{findingbox}[Finding \#3]
Compression resilience varies by task. Code summarization and clone detection tolerate higher compression ratios, with some models maintaining performance at 4×--8×. Code completion and question answering are more sensitive, with most models showing degradation beyond 2×--4× compression.
\end{findingbox}

\input{tables/rq3.tex}

\subsubsection{Compression Effects Across Models}

Model capability strongly influences compression resilience.
Across all four tasks, the Gemini-3 family (Gemini-3-Flash and Gemini-3-Pro) demonstrates remarkable compression resilience, with no significant degradation and even significant improvements in code completion and question answering at 8×.
In code completion, Gemini-3-Pro achieves ES 58.0 at 8× compared to 55.8 with raw text.
In code question answering, Gemini-3-Pro reaches 79.5\% accuracy at 8× versus 74.8\% with raw text.

In contrast, models with weaker visual understanding capabilities show more pronounced degradation.
GLM-4.6v's accuracy in code question answering drops significantly from 72.6\% at 1× to 39.7\% at 8× ($p<0.01$).
GPT-5-mini and GPT-5.1 exhibit moderate resilience in some tasks but inconsistent performance in others.

These observations suggest that compression resilience correlates with overall model capability in code image understanding, and that state-of-the-art models are better equipped to handle compressed visual inputs.

\begin{findingbox}[Finding \#4]
Compression resilience also varies by model. Gemini-3-Pro maintains or improves performance at 8× compression (only 12.5\% of text tokens), while models with weaker visual capabilities show pronounced degradation at higher ratios.
\end{findingbox}

\subsection{RQ3: Can Visual Enhancements Improve Code Image Understanding?}


In RQ1--RQ2, we used plain rendering (the default configuration). Here, we investigate whether enhanced rendering strategies---bold and syntax highlighting---can further improve model performance.
To assess statistical significance, we apply the Wilcoxon signed-rank test~\cite{wilcoxon1945individual} to compare Plain rendering against Bold and Highlight variants, with the null hypothesis that there is no significant difference.
The results are presented in Table~\ref{tab:rq3_all}.

\subsubsection{Enhancement Effectiveness at Low-to-Moderate Compression}

Visual enhancements---including syntax highlighting and bold rendering---can significantly improve LLMs' code image understanding, particularly at compression ratios of 1×--4× where the visual signal remains legible.

In \textit{code completion}, both bold rendering and syntax highlighting provide significant improvements.
At 1× compression, GLM-4.6v improves significantly from ES 50.8 (plain) to 53.2 with highlighting ($p<0.01$), and GPT-5.1 benefits significantly from bold rendering (ES: 50.1 vs. 47.9, $p<0.01$).
The Gemini family demonstrates particularly strong responsiveness: Gemini-3-Flash achieves significant improvements across both strategies at 1×--2× compression ($p<0.05$).

In \textit{clone detection}, GPT-5.1 shows significant improvement with bold rendering at 4× compression (F1: 64.6 vs. 58.2, +11\%, $p<0.01$), and Gemini-2.5-Pro benefits significantly from both strategies at moderate compression levels.
For \textit{code question answering}, Gemini-3-Flash achieves 76.8\% accuracy with bold rendering at 1× compression, compared to 74.8\% with plain rendering ($p<0.01$).

\begin{findingbox}[Finding \#5]
Visual enhancements are most effective at 1×--4× compression, with state-of-the-art models like the Gemini family showing consistent improvements of 2--5\% across tasks.
\end{findingbox}

\subsubsection{Diminishing Returns at High Compression}

At 8× compression, visual enhancements generally offer limited additional benefit, as reduced resolution obscures the visual distinctions they introduce.
However, some model-task combinations still show significant improvements: Gemini-3-Pro with bold rendering in code summarization ($p<0.01$) and Gemini-3-Flash with bold rendering in clone detection ($p<0.01$).
Bold rendering can even introduce slight degradation at extreme compression for some models, as thicker strokes may reduce character distinguishability.
The varying effectiveness across models suggests that visual enhancement responsiveness depends on model-specific factors, representing an optimization opportunity for future work.

These findings suggest that enhancement strategies should be adapted to compression level---meaningful at moderate compression, but potentially unnecessary at high compression ratios.

\begin{findingbox}[Finding \#6]
At 8× compression, visual enhancements provide diminishing returns. Adaptive rendering strategies that adjust enhancement based on target compression ratio represent a promising direction for future optimization.
\end{findingbox}


\subsection{RQ4: Can Code Image Understanding Generalize to Other Languages?}

To validate the generalizability of our findings from RQ1--RQ3 beyond Python, we extend key experiments to Java---a language with fundamentally different syntactic characteristics (explicit braces vs. whitespace indentation). We evaluate code completion and clone detection using Java benchmarks detailed in Section~\ref{sec:bm}.
Following the same statistical testing protocol as RQ1--RQ3, we apply the Wilcoxon signed-rank test~\cite{wilcoxon1945individual} to assess significance.
The results are presented in Table~\ref{tab:rq4_java}.

\input{tables/rq4.tex}

The fundamental patterns observed in Python experiments replicate consistently in Java.
In \textit{code completion}, the Gemini family significantly outperforms raw text across all compression levels ($p<0.01$), demonstrating strong visual code understanding.
In \textit{clone detection}, visual inputs provide significant improvements across multiple models ($p<0.01$).
Qwen-3-VL shows particularly large gains under compression (F1: 24.2 $\rightarrow$ 53.0 at 8×, +119\%, $p<0.01$), suggesting that compression may blur syntactic details and encourage the model to focus on higher-level semantic patterns rather than surface-level differences.
The compression resilience patterns also hold: models that performed well under compression in Python maintain their relative advantages in Java.

\begin{findingbox}[Finding \#7]
Core findings generalize from Python to Java. The patterns of code image understanding---including model-specific strengths and compression resilience---remain consistent across languages.
\end{findingbox}

\subsection{RQ5: How Does Visual Compression Degrade the Information in Code?}
\label{sec:rq5}

RQ1--RQ4 revealed that compression affects different tasks differently---summarization and clone detection remain resilient while code completion shows more variation.
This raises a key question: \textit{how} does compression degrade the information in code, and why does this impact tasks differently?
To answer this, we design a code reconstruction experiment that directly measures information preservation, where models are instructed to transcribe the code from compressed code images.
The errors between the original code and reconstructed code thereby reveal what visual information is lost.

To ensure uncontaminated evaluation, we utilize the GitHub REST API~\cite{github_rest_api} to fetch fresh Python repositories created strictly after August 1, 2025 (after the knowledge cutoff of all studied models), filtering for repositories with 10+ stars and file lengths between 50--120 lines to ensure code quality while maintaining suitability for code image generation.
To ensure diversity, we selected top 100 repositories (excluding those used in Code Question Answering) and randomly selected one code snippet per repository that meets the target length criteria, resulting in 100 code snippets with an average length of 473.1 tokens. 
We evaluate all seven models across four compression ratios (1×, 2×, 4×, 8×), using a strict OCR prompt: ``\textit{Transcribe the code in the image exactly}''.
Reconstruction quality is measured using Character Error Rate (CER)~\cite{thennal2025advocating}, CodeBLEU~\cite{ren2020codebleu}, and Exact Match (EM).

\begin{figure}[t]
    \centering
    \includegraphics[width=\linewidth]{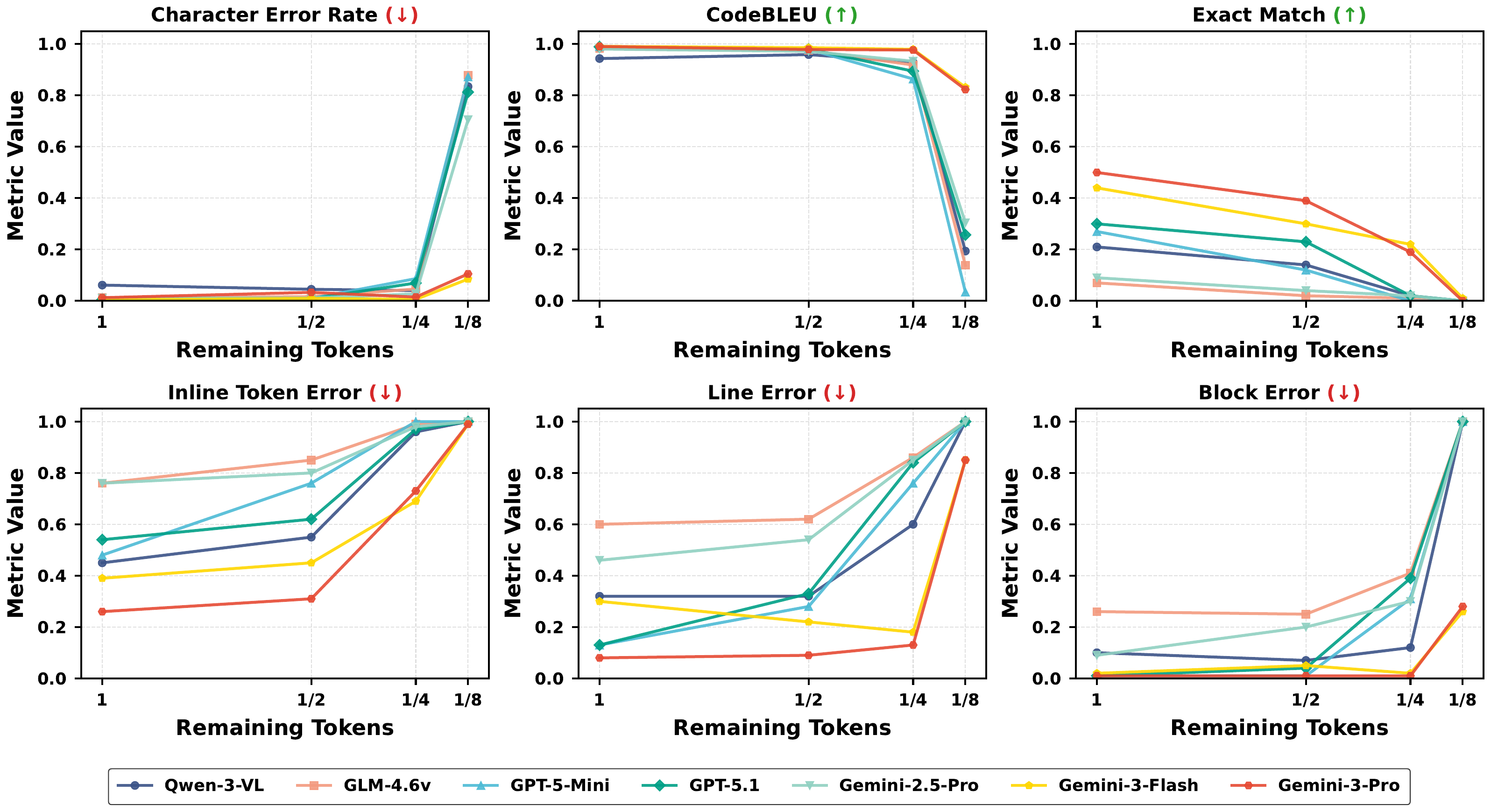}
    \vspace{-0.25cm}
    \caption{Code Reconstruction Performance across Different Remaining Token Ratios.}
    \vspace{-0.5cm}
    \label{Fig:RQ5_OCR}
\end{figure}

We further categorize errors using a rule-based three-level taxonomy: \textit{Token Error} (non whitespace tokens that differ from the ground truth), \textit{Line Error} (a line where $\geq$50\% of tokens differ), and \textit{Block Error} (three or more consecutive Line Errors). We quantify these errors by measuring their Prevalence: the percentage of samples containing at least one instance of a specific error type.
The results are presented in Figure~\ref{Fig:RQ5_OCR}.


\subsubsection{Visual Information Loss Patterns}
Compression degrades code reconstruction quality in predictable patterns, with model capability determining resilience.
At 1× compression, Gemini-3-Pro achieves the highest Exact Match and lowest CER, followed by Gemini-3-Flash and GPT-5.1.

Under compression, we observe two distinct patterns.
The Gemini-3 family demonstrates ``graceful degradation,'' maintaining high CodeBLEU even at 8× compression---likely due to training objectives that emphasize visual document understanding~\citep{gemini3pro_model_card_2025}.
Other models exhibit a ``performance cliff'' pattern, maintaining reasonable accuracy until 4× compression before rapid decline at 8×.
These reconstruction quality differences directly predict downstream task performance: models with graceful degradation (Gemini-3) excel across all tasks in RQ1--RQ4, while models with performance cliffs show task-dependent results.

\subsubsection{Error Type Analysis}
\label{subsec:error_analysis}

The results reveal a clear degradation hierarchy.
\textit{Token Errors} emerge first---even at 1× compression, most models show token errors (e.g., confusing \texttt{1} vs \texttt{l}, \texttt{0} vs \texttt{O}, missing punctuation), with error rates increasing under compression.
\textit{Line Errors} remain relatively stable from 1× to 4×, then surge dramatically at 8× for most models.
\textit{Block Errors} dominate at aggressive compression for weaker models, indicating that they begin ``hallucinating'' code rather than transcribing.
Notably, the Gemini-3 family maintains low block error rates even at 8× compression, which directly explains their consistent performance on downstream tasks across all compression levels observed in RQ1--RQ4.

The error hierarchy directly explains the task-dependent patterns observed in RQ1--RQ4.
First, \textit{some downstream tasks do not require perfect reconstruction}: even when Token Error prevalence is high, models can still achieve competitive performance on summarization and clone detection, because these tasks rely on high-level semantic patterns rather than character-level precision.
This explains the apparent paradox where models with substantial reconstruction errors still perform well on downstream tasks---the OCR task demands exact transcription, while downstream tasks only require sufficient semantic understanding.
Second, \textit{detail-sensitive tasks degrade with error accumulation}: code completion and question answering benefit from low error rates, explaining why models with graceful degradation (Gemini-3) excel while others show more variation.
This causal link---from visual information loss patterns to downstream task performance---validates our reconstruction analysis as a diagnostic tool for understanding code image understanding capabilities.

\begin{findingbox}[Finding \#8]
Information degradation follows a predictable hierarchy: Token Errors emerge first (1×--2×), Line Errors at moderate compression (2×--4×), and Block Errors at high compression (4×--8×) for most models. However, the Gemini-3 family maintains low block error rates even at 8× compression, explaining their stable downstream performance across all compression levels.
\end{findingbox}





\section{Discussion}

\subsection{Inference Latency}

\begin{wrapfigure}{r}{0.4\textwidth} 
  \centering
  \vspace{-15pt} 
  \includegraphics[width=\linewidth]{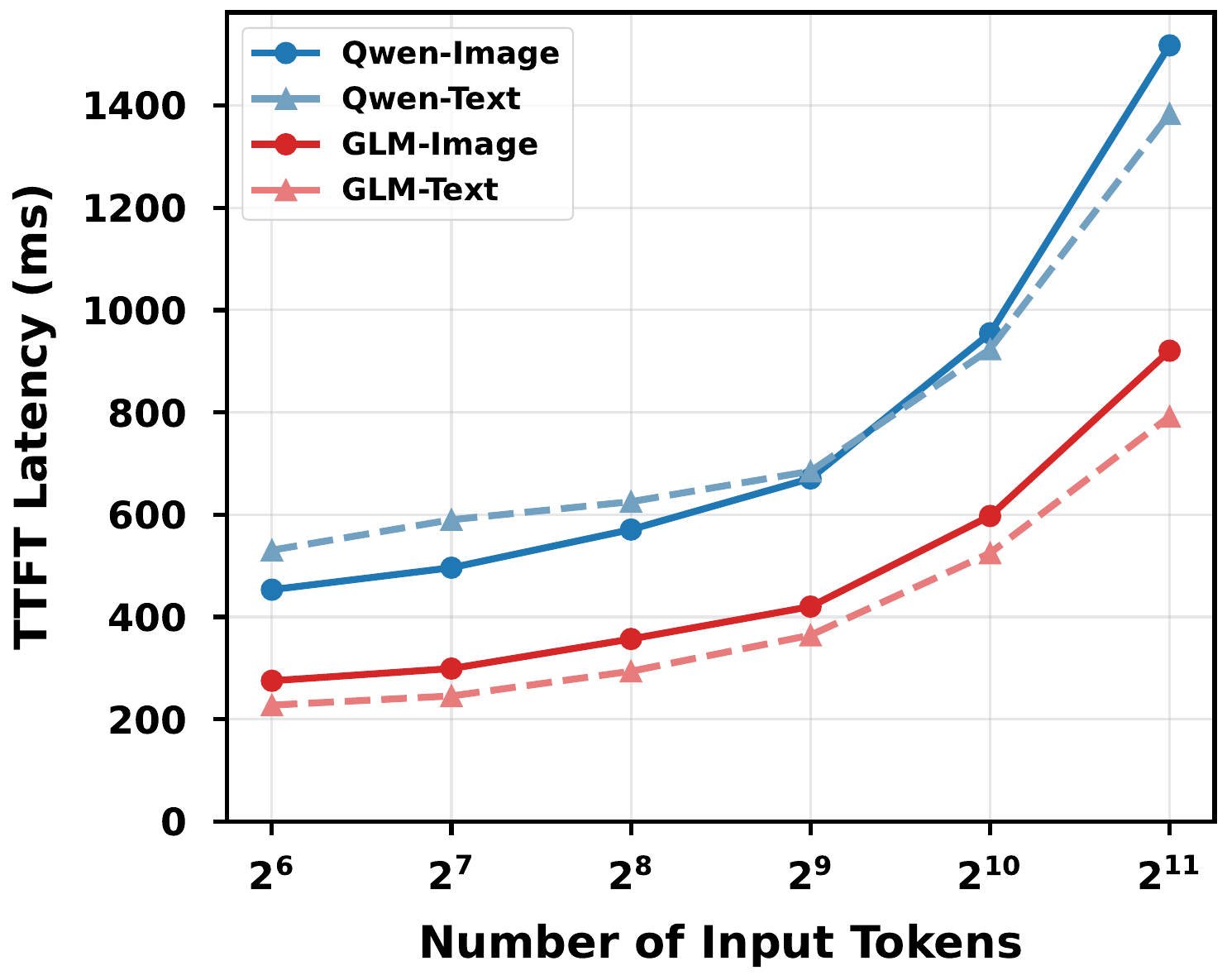}
  \caption{Time-to-First-Token (TTFT) Comparison: Text vs. Image Inputs}
  \label{fig:ttft-merged}
  \vspace{-5pt} 
\end{wrapfigure}

A key question for practical deployment is whether visual code processing introduces prohibitive latency overhead compared to text-based approaches.
While commercial API providers typically charge the same rate for visual and text tokens~\cite{openai_api_pricing_2025,gemini_api_pricing_2025}, the actual computational cost may differ due to the additional visual encoder and alignment stages (\Cref{fig:mllm}).
Since API latency is heavily influenced by network conditions and server load, we benchmark locally on two open-weight MLLMs: Qwen-3-VL (235B)\footnote{\url{https://huggingface.co/Qwen/Qwen3-VL-235B-A22B-Instruct}} and GLM-4.6v (108B)\footnote{\url{https://huggingface.co/zai-org/GLM-4.6V}}, measuring Time to First Token (TTFT)---the latency from input submission to the first generated token, encompassing both prefill and initial decoding~\citep{agrawal2024metron}, which is a widely used measure that reflects the perceived responsiveness for interactive developer tools~\citep{agarwal2023llm,zhong2024distservedisaggregatingprefilldecoding,agrawal2025evaluatingperformancellminference}. 
Experiments are conducted on a machine with 8×NVIDIA A100-80G GPUs, dual-socket AMD EPYC 7763 CPUs (128 cores), and 1.8TB system memory.
We use PyTorch~\citep{paszke2019pytorch} with the HuggingFace Transformers library~\citep{wolf2020transformers} for inference.
Each measurement consists of 2 warmup iterations followed by 10 timed iterations, with the average execution time reported.

As shown in Figure~\ref{fig:ttft-merged}, the latency curves for images and text are comparable at identical token scales, indicating that visual encoding introduces minimal overhead.
With per-token latency parity established, the 2×--4× compression ratios demonstrated in RQ2--3 directly translate to equivalent inference speedup—processing 4× compressed images is approximately 4× faster than processing raw text.
Interestingly, Qwen-3-VL shows slightly \textit{lower} latency for images than text at small token counts ($<2^9$), likely due to the parallel processing efficiency of vision encoders on small inputs compared to sequential text tokenization overhead~\citep{wolf2020transformers}.
These results suggest that code image understanding is not only viable in terms of task performance but also practically deployable without latency penalties, paving the way for ``vision-first'' code intelligence systems.

\subsection{Threats to Validity}

\textbf{Internal Validity.}
The primary internal threat is \textit{data contamination}---commercial MLLMs may have seen benchmark data during pre-training~\citep{fang2025lastingbench}. We address this through multiple strategies: (1) No Context baselines that remove retrieved context while preserving task-specific inputs, allowing us to isolate the contribution of visual context; (2) constructing our CodeQA dataset exclusively from GitHub repositories created after August 2025, well beyond model training cutoffs; and (3) crawling 100 fresh repositories for RQ5's code reconstruction experiments (Section~\ref{sec:bm}). To ensure \textit{annotation quality}, three independent researchers validated each CodeQA question-answer pair, with unanimous agreement required for inclusion; samples with any disagreement were discarded. For \textit{statistical reliability}, we repeat all experiments five times and apply Wilcoxon signed-rank tests to assess significance (Section~\ref{sec:implementation}).

\textbf{External Validity.}
For \textit{programming language coverage}, while our primary experiments focus on Python, we replicate key findings on Java in RQ4, demonstrating consistent performance patterns across languages with different syntax characteristics. For \textit{model coverage}, we evaluate seven representative MLLMs from diverse families---including open-weight models (Qwen-3-VL, GLM-4.6v) and proprietary systems (GPT-5-mini, GPT-5.1, Gemini-2.5-Pro, Gemini-3-Flash, Gemini-3-Pro)---to capture the spectrum of current capabilities (Section~\ref{sec:llms}). For \textit{rendering configuration}, rather than exploring exhaustive visual parameters, we adopt VSCode's default syntax highlighting theme---a widely-used IDE configuration---to maximize practical relevance (Section~\ref{sec:visual_processing}).

\section{\ourmethod: Code Transformation Tool}
\label{sec:tool}

\begin{wrapfigure}{r}{0.3\textwidth}
  \centering
  \vspace{-15pt}
  \includegraphics[width=0.9\linewidth]{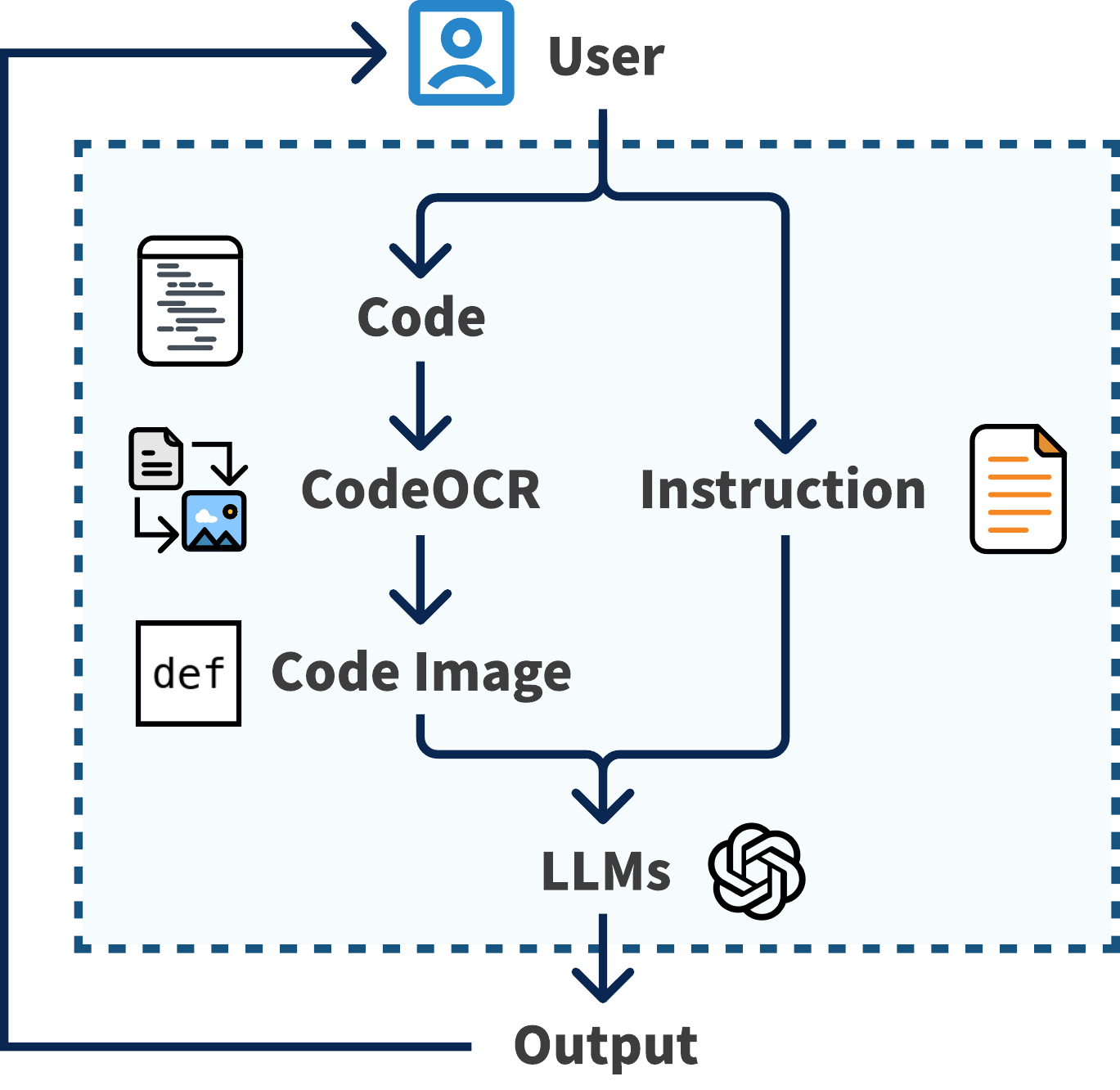}
  \caption{CodeOCR Workflow}
  \label{fig:CodeOCR_Workflow}
  \vspace{-5pt}
\end{wrapfigure}

Our experiments reveal that visual code representation offers a promising paradigm for MLLM-based code understanding, achieving comparable or improved performance at significant compression ratios. Building on these findings, we developed \ourmethod, a practical middleware for rendering source code into images with configurable visual enhancements and compression ratios.

\noindent \textbf{Workflow.}
As illustrated in Figure~\ref{fig:CodeOCR_Workflow}, users provide code and instructions as input; \ourmethod~renders the code into a compact image, which is then passed to the MLLM along with the instructions, and the output is returned to the user. Internally, the transformation comprises two stages: (1) \textit{Visual Rendering} converts source code into syntax-highlighted images, and (2) \textit{Dynamic Compression} adjusts resolution to achieve target compression ratios based on user-specified token budgets. The tool leverages \texttt{Pygments}~\cite{pygments} for syntax analysis and \texttt{Pillow}~\cite{pillow} for image rendering, currently supporting six languages (Python, Java, JavaScript, C/C++, Go, TypeScript) with native extensibility to 500+ languages via Pygments' lexer ecosystem.

\noindent \textbf{Usage Scenarios.}
\ourmethod~serves as an efficient middleware for both LLM service providers and end-users. By converting code into compact images, it significantly reduces the computational overhead and financial costs associated with API usage. This transformation is applicable to code of any scale---from individual functions to entire projects---enabling users to trade visual fidelity for token savings based on their specific needs.

\noindent \textbf{Performance Testing.}
We evaluated the middleware's efficiency using over 1,000 samples across four benchmarks. Performance tests demonstrate that \ourmethod~achieves a high transformation throughput of 6.9k token/s, making it sufficiently fast for real-time applications or on-the-fly processing in IDE plugins. We further validated the tool's reliability by confirming 100\% consistency in token estimation and compression ratio accuracy across repeated runs. The source code and reproduction scripts are available online~\cite{codeocr_repo}.

\section{Related Work}
\textbf{Large Language Models for Code.}
Recent years have witnessed rapid advancement in LLMs for code~\citep{fan2023large,jiang2024survey,zhang2024unifying,wang2025uncertainty,shi2026swe,chen2026classeval}. Starting from Codex~\citep{chen2021codex}, a series of code LLMs including Code Llama~\citep{roziere2023codellama}, StarCoder~\citep{lozhkov2024starcoder2}, DeepSeek-Coder~\citep{guo2024deepseek,zhu2024deepseekcoderv2}, and Qwen2.5-Coder~\citep{hui2024qwen25coder} have achieved strong performance across diverse tasks such as code generation~\citep{hu2026line,zeng2026glimprouter,wang2026effiskill}, repair~\citep{muennighoff2023octopack,shi2024code,li2025swe,chen2025swe,chen2025unveiling,chen2025evaluating,wang2026swe,wang2026swe2,hu2026zeroshot}, translation~\citep{khan2024xcodeeval,ahmad2023avatar,wang2025evoc2rust,hu2025flowmaltrans}, and reasoning~\citep{gu2024cruxeval,zeng2025pruning,peng2025swe}. 
While these models have achieved remarkable success, they process code as linear token sequences, facing scalability challenges as context length grows~\cite{guo2023longcoder,bogomolovLongCodeArena2024}. Text-based compression methods~\citep{zhang2022dietcode,wang2024slimcode,shi2025longcodezip,yang2024shortendoc,pan2025hiddencost,sun2024ai,fang2025attentionrag,wang2026context} alleviate this through selective token retention, i.e., each token is either kept or dropped. This inevitably leads to a certain degree of information loss, and those kept key tokens cannot be further compressed~\citep{wang2026swe,shi2025longcodezip,sun2025token}. Our work explores a complementary paradigm: representing code as images enables continuous compression via resolution scaling rather than discrete selection. Empirically, we find MLLMs achieve comparable performance at up to 8× compression, highlighting visual representation of code as a promising research direction.

\noindent\textbf{Visual Document Understanding.}
OCR and visual document understanding have evolved from traditional digitization~\cite{10.1145/3606692,4376991,jaderberg2014readingtextwildconvolutional,baek2019wrongscenetextrecognition,long2020textsnakeflexiblerepresentationdetecting,katti2018chargridunderstanding2ddocuments,Xu_2020,rausch2021docparserhierarchicalstructureparsing} to end-to-end neural approaches. Early systems like TrOCR~\cite{li2022trocrtransformerbasedopticalcharacter} and Nougat~\cite{blecher2023nougat} demonstrated direct transcription without separate detection stages, while GOT-OCR2.0~\cite{wei2024gotorc} enhanced structure recovery for charts and tables.
General-purpose MLLMs~\cite{openai2023gpt4,team2023gemini,liu2024llava,chen2024internvlscalingvisionfoundation,kim2022ocrfreedocumentunderstandingtransformer,lee2023pix2structscreenshotparsingpretraining,bai2023qwenvl} have advanced high-resolution visual understanding, with specialized models for document comprehension~\cite{wang2024docllm,hu2024mplugdocowl,chen2025progressive} and GUI understanding~\cite{you2024ferretui,baechler2024screenai}. DeepSeek-OCR~\cite{wei2025deepseekocr} introduced optical compression for documents, achieving up to 20× ratios.
However, these works focus on natural documents or UI screenshots, where visual layouts are loosely structured. Code presents unique challenges with dense symbolic content and strict syntactic constraints~\cite{buse2010learning,storey2006theories}. Our study systematically evaluates how MLLMs handle code-specific visual fidelity under compression---revealing task-dependent resilience patterns not explored in prior OCR research.

\section{Conclusion and Future Directions}

This paper presents the first comprehensive empirical study exploring visual code representation as a new paradigm for code understanding.
Through systematic evaluation of state-of-the-art MLLMs across four representative tasks, we provide empirical evidence that this paradigm is both viable and practically beneficial.
Our findings offer actionable insights for future research and practice.
First, we observe that image compression can achieve competitive or even superior performance while using only 25\% or fewer tokens---this suggests that for practitioners, visual representation can substantially reduce API costs without sacrificing quality, and motivates the design of code-specific compression techniques.
Second, we find that syntax highlighting improves model robustness, indicating opportunities for task-adaptive rendering strategies.
Third, we identify significant OCR capability gaps across models, pointing to the need for code-specific visual pre-training.
These findings establish visual code representation as a promising research direction and motivate future work on task-adaptive rendering, aggressive compression techniques, and code-specialized multimodal models.

\section*{Acknowledgments}
This paper is supported by the National Key Research and Development Program of China (Grant No. 2023YFB4503802), the National Natural Science Foundation of China (Grant No. 62402183), the Natural Science Foundation of Shanghai (Grant No. 25ZR1401175), and the National Research Foundation, Singapore, under its Investigatorship Grant (NRF-NRFI08-2022-0002). Any opinions, findings and conclusions or recommendations expressed in this material are those of the author(s) and do not reflect the views of National Research Foundation, Singapore.

\section*{Data Availability}

The prompts, source code, datasets, and reproduction scripts are publicly available in our replication package: \url{https://github.com/YerbaPage/CodeOCR}.

\bibliographystyle{ACM-Reference-Format}
\bibliography{ref}

\appendix

\end{document}

%% file: tables/rq1.tex
\begin{table}[t]
    \centering
    \caption{Overall Performance of MLLMs on Downstream Tasks with Different Inputs.}
    \vspace{-0.25cm}
    \label{tab:rq1_overall}
    \resizebox{\textwidth}{!}{
    \begin{tabular}{l ccc ccc ccc ccc}
    \toprule
    \multirow{3}{*}{\textbf{Model}} & \multicolumn{3}{c}{\textbf{Code Summarization}} & \multicolumn{3}{c}{\textbf{Code Completion}} & \multicolumn{3}{c}{\textbf{Code Clone Detection}} & \multicolumn{3}{c}{\textbf{Code Question Answering}} \\
     & \multicolumn{3}{c}{\small{CompScore (\%)}} & \multicolumn{3}{c}{\small{ES / EM (\%)}} & \multicolumn{3}{c}{\small{ACC / F1 (\%)}} & \multicolumn{3}{c}{\small{ACC (\%)}} \\
    \cmidrule(lr){2-4} \cmidrule(lr){5-7} \cmidrule(lr){8-10} \cmidrule(lr){11-13}
     & \textbf{NoCtx} & \textbf{Text} & \textbf{Image} & \textbf{NoCtx} & \textbf{Text} & \textbf{Image} & \textbf{NoCtx} & \textbf{Text} & \textbf{Image} & \textbf{NoCtx} & \textbf{Text} & \textbf{Image} \\
    \midrule
    Qwen-3-VL & -- & 56.6 & 56.4 & 45.0/12.8 & 49.7/21.6 & 35.5$^{**}$/8.0$^{**}$ & -- & 67.8/52.2 & 67.2/51.2 & 45.6 & 84.0 & 58.1$^{**}$ \\
     & & \scriptsize{$\pm$2.8} & \scriptsize{$\pm$4.1} & \scriptsize{$\pm$0.4/0.6} & \scriptsize{$\pm$0.2/0.9} & \scriptsize{$\pm$0.4/1.1} & & \scriptsize{$\pm$0.4/0.7} & \scriptsize{$\pm$0.4/0.9} & \scriptsize{$\pm$0.2} & \scriptsize{$\pm$0.0} & \scriptsize{$\pm$1.2} \\
    \midrule
    GLM-4.6v & -- & 55.4 & 54.6 & 39.9/9.5 & 49.8/21.0 & 50.8/17.2$^{*}$ & -- & 81.6/78.4 & 69.6$^{**}$/58.2$^{**}$ & 37.2 & 78.5 & 72.6$^{*}$ \\
     & & \scriptsize{$\pm$2.0} & \scriptsize{$\pm$1.6} & \scriptsize{$\pm$0.8/0.3} & \scriptsize{$\pm$0.4/1.2} & \scriptsize{$\pm$0.7/0.5} & & \scriptsize{$\pm$0.5/1.2} & \scriptsize{$\pm$0.5/1.9} & \scriptsize{$\pm$2.5} & \scriptsize{$\pm$1.6} & \scriptsize{$\pm$1.3} \\
    \midrule
    GPT-5-mini & -- & 57.1 & 56.5 & 42.6/10.7 & 51.3/25.3 & 51.6/24.7 & -- & 59.4/33.2 & \cellcolor{green!15}64.8$^{**}$/47.0$^{**}$ & 46.3 & 82.0 & 77.5$^{*}$ \\
     & & \scriptsize{$\pm$0.6} & \scriptsize{$\pm$2.1} & \scriptsize{$\pm$0.8/1.0} & \scriptsize{$\pm$1.1/1.9} & \scriptsize{$\pm$1.8/1.7} & & \scriptsize{$\pm$0.5/1.3} & \cellcolor{green!15}\scriptsize{$\pm$0.4/1.4} & \scriptsize{$\pm$1.9} & \scriptsize{$\pm$1.1} & \scriptsize{$\pm$1.4} \\
    \midrule
    GPT-5.1 & -- & 56.6 & 55.9 & 41.8/10.7 & 51.6/24.6 & 47.9$^{*}$/18.6$^{**}$ & -- & 65.8/46.8 & \cellcolor{green!15}71.6$^{**}$/62.4$^{**}$ & 44.7 & 79.3 & 68.6$^{**}$ \\
     & & \scriptsize{$\pm$0.4} & \scriptsize{$\pm$1.6} & \scriptsize{$\pm$0.4/1.3} & \scriptsize{$\pm$0.9/1.2} & \scriptsize{$\pm$1.4/2.0} & & \scriptsize{$\pm$0.4/1.0} & \cellcolor{green!15}\scriptsize{$\pm$0.5/1.2} & \scriptsize{$\pm$1.4} & \scriptsize{$\pm$1.2} & \scriptsize{$\pm$1.5} \\
    \midrule
    Gemini-2.5-Pro & -- & 54.5 & \cellcolor{green!5}55.2 & 46.7/15.5 & 54.9/28.8 & 54.5/25.7$^{*}$ & -- & 65.4/46.4 & \cellcolor{green!10}67.0/50.6 & 40.2 & 82.2 & 71.2$^{**}$ \\
     & & \scriptsize{$\pm$0.7} & \cellcolor{green!5}\scriptsize{$\pm$2.1} & \scriptsize{$\pm$0.8/0.6} & \scriptsize{$\pm$0.8/0.5} & \scriptsize{$\pm$1.2/1.8} & & \scriptsize{$\pm$1.1/2.5} & \cellcolor{green!10}\scriptsize{$\pm$1.2/2.6} & \scriptsize{$\pm$3.0} & \scriptsize{$\pm$1.4} & \scriptsize{$\pm$0.6} \\
    \midrule
    Gemini-3-Flash & -- & 55.2 & \cellcolor{green!5}55.5 & 49.7/15.8 & 55.1/26.5 & \cellcolor{green!10}57.1$^{*}$/29.2$^{*}$ & -- & 70.0/59.4 & 67.8/55.4$^{*}$ & 47.9 & 73.4 & \cellcolor{green!5}74.8$^{*}$ \\
     & & \scriptsize{$\pm$1.0} & \cellcolor{green!3}\scriptsize{$\pm$1.8} & \scriptsize{$\pm$0.4/0.5} & \scriptsize{$\pm$0.4/0.8} & \cellcolor{green!10}\scriptsize{$\pm$0.6/1.0} & & \scriptsize{$\pm$1.1/1.5} & \scriptsize{$\pm$1.2/1.4} & \scriptsize{$\pm$2.6} & \scriptsize{$\pm$0.4} & \cellcolor{green!5}\scriptsize{$\pm$1.2} \\
    \midrule
    Gemini-3-Pro & -- & 56.0 & \cellcolor{green!5}56.8 & 50.3/16.2 & 55.8/27.6 & \cellcolor{green!10}57.7$^{*}$/29.2$^{*}$ & -- & 71.0/60.8 & 70.2/58.8 & 46.8 & 74.8 & \cellcolor{green!10}77.2$^{*}$ \\
     & & \scriptsize{$\pm$1.2} & \cellcolor{green!5}\scriptsize{$\pm$1.5} & \scriptsize{$\pm$0.4/0.7} & \scriptsize{$\pm$0.5/0.8} & \cellcolor{green!10}\scriptsize{$\pm$0.6/0.9} & & \scriptsize{$\pm$0.9/1.8} & \scriptsize{$\pm$1.0/1.6} & \scriptsize{$\pm$2.2} & \scriptsize{$\pm$1.2} & \cellcolor{green!10}\scriptsize{$\pm$1.4} \\
    \bottomrule
    \end{tabular}
    }
    \vspace{2mm}
    {\scriptsize \makebox[\textwidth][r]{\colorbox{green!10}{Green}: Image outperforms Text.\quad\quad$^{*}$: $p$-value $< 0.05$\quad\quad$^{**}$: $p$-value $< 0.01$}}
    \vspace{-0.5cm}
\end{table}

%% file: tables/rq3.tex
\begin{table}[ht]
\centering
\caption{Impact of Visual Rendering Strategies (Plain, Bold, Highlight) on Code Understanding Tasks.}
\vspace{-0.25cm}
\label{tab:rq3_all}
\resizebox{1\textwidth}{!}{
\setlength{\tabcolsep}{4pt}
\begin{tabular}{l ccc ccc ccc ccc}
\toprule
\multirow{3}{*}{\textbf{Model}} &
\multicolumn{3}{c}{\textbf{Image (1x)}} &
\multicolumn{3}{c}{\textbf{Image (2x)}} &
\multicolumn{3}{c}{\textbf{Image (4x)}} &
\multicolumn{3}{c}{\textbf{Image (8x)}} \\
 &
\multicolumn{3}{c}{\small{100\% tokens}} &
\multicolumn{3}{c}{\small{50\% tokens}} &
\multicolumn{3}{c}{\small{25\% tokens}} &
\multicolumn{3}{c}{\small{12.5\% tokens}} \\
\cmidrule(lr){2-4} \cmidrule(lr){5-7} \cmidrule(lr){8-10} \cmidrule(lr){11-13}
 &
\textbf{Plain} & \textbf{Bold} & \textbf{Highlight} &
\textbf{Plain} & \textbf{Bold} & \textbf{Highlight} &
\textbf{Plain} & \textbf{Bold} & \textbf{Highlight} &
\textbf{Plain} & \textbf{Bold} & \textbf{Highlight} \\
\midrule
\multicolumn{13}{c}{\textbf{Code Completion (ES/EM, \%)}} \\
\midrule
Qwen-3-VL & 35.5/8.0 & 34.5/9.2 & \cellcolor{green!5}36.2/10.5 & 35.7/9.2 & \cellcolor{green!2}36.0/9.9 & 33.8/8.2 & 38.5/9.7 & \cellcolor{green!2}38.8/10.1 & 37.7/10.7 & 41.1/12.8 & \cellcolor{green!2}41.1/12.4 & 41.0/11.9 \\
 & \scriptsize{$\pm$0.4/1.1} & \scriptsize{$\pm$0.7/0.9} & \cellcolor{green!5}\scriptsize{$\pm$0.7/0.9} & \scriptsize{$\pm$1.0/0.8} & \cellcolor{green!2}\scriptsize{$\pm$0.5/0.5} & \scriptsize{$\pm$0.6/0.7} & \scriptsize{$\pm$0.8/0.7} & \cellcolor{green!2}\scriptsize{$\pm$0.3/0.4} & \scriptsize{$\pm$0.5/0.8} & \scriptsize{$\pm$0.4/0.8} & \cellcolor{green!2}\scriptsize{$\pm$0.6/0.5} & \scriptsize{$\pm$0.3/0.7} \\
GLM-4.6v & 50.8/17.2 & \cellcolor{green!10}52.3$^{*}$/18.0 & \cellcolor{green!15}53.2$^{**}$/20.7$^{**}$ & 46.3/11.8 & \cellcolor{green!5}47.3/12.5 & \cellcolor{green!8}47.7$^{*}$/13.6$^{*}$ & 44.1/10.5 & 44.1/10.2 & \cellcolor{green!10}46.0$^{*}$/12.0$^{*}$ & 42.9/8.9 & 42.8/8.9 & \cellcolor{green!5}44.0$^{*}$/10.2$^{*}$ \\
 & \scriptsize{$\pm$0.7/0.5} & \cellcolor{green!10}\scriptsize{$\pm$1.0/1.3} & \cellcolor{green!15}\scriptsize{$\pm$1.3/1.9} & \scriptsize{$\pm$0.4/0.9} & \cellcolor{green!5}\scriptsize{$\pm$0.9/0.7} & \cellcolor{green!8}\scriptsize{$\pm$0.5/0.6} & \scriptsize{$\pm$0.7/1.2} & \scriptsize{$\pm$0.8/1.2} & \cellcolor{green!10}\scriptsize{$\pm$0.6/0.6} & \scriptsize{$\pm$0.3/0.2} & \scriptsize{$\pm$0.6/0.8} & \cellcolor{green!5}\scriptsize{$\pm$0.6/1.4} \\
GPT-5-mini & 51.6/24.7 & \cellcolor{green!5}52.5/25.2 & 49.1/21.9 & 45.6/16.7 & \cellcolor{green!10}47.3$^{*}$/17.3 & \cellcolor{green!10}47.1$^{*}$/17.7 & 44.4/14.1 & 43.7/12.8 & 44.2/14.9 & 43.3/12.3 & 42.6/11.6 & \cellcolor{green!5}44.1/13.5$^{*}$ \\
 & \scriptsize{$\pm$1.8/1.7} & \cellcolor{green!5}\scriptsize{$\pm$1.1/1.0} & \scriptsize{$\pm$1.6/1.5} & \scriptsize{$\pm$1.0/1.2} & \cellcolor{green!10}\scriptsize{$\pm$1.8/1.8} & \cellcolor{green!10}\scriptsize{$\pm$1.3/1.3} & \scriptsize{$\pm$1.7/1.7} & \scriptsize{$\pm$0.5/0.9} & \scriptsize{$\pm$0.8/0.7} & \scriptsize{$\pm$1.0/1.4} & \scriptsize{$\pm$1.2/1.7} & \cellcolor{green!5}\scriptsize{$\pm$1.0/1.1} \\
GPT-5.1 & 47.9/18.6 & \cellcolor{green!20}50.1$^{**}$/18.3 & \cellcolor{green!5}48.0/19.7 & 46.5/13.1 & \cellcolor{green!5}47.0/14.5$^{*}$ & \cellcolor{green!10}47.4/16.0$^{**}$ & 46.6/13.8 & \cellcolor{green!5}46.8/12.1 & \cellcolor{green!10}47.4/16.0$^{*}$ & 45.5/13.1 & 45.0/12.8 & \cellcolor{green!10}46.3/13.8 \\
 & \scriptsize{$\pm$1.4/2.0} & \cellcolor{green!20}\scriptsize{$\pm$1.4/1.4} & \cellcolor{green!5}\scriptsize{$\pm$1.7/1.3} & \scriptsize{$\pm$1.7/1.8} & \cellcolor{green!5}\scriptsize{$\pm$0.6/0.6} & \cellcolor{green!10}\scriptsize{$\pm$1.0/0.8} & \scriptsize{$\pm$0.8/1.8} & \cellcolor{green!5}\scriptsize{$\pm$0.9/1.2} & \cellcolor{green!10}\scriptsize{$\pm$0.2/1.3} & \scriptsize{$\pm$0.3/0.4} & \scriptsize{$\pm$1.1/0.9} & \cellcolor{green!10}\scriptsize{$\pm$1.6/1.0} \\
Gemini-2.5-Pro & 54.5/25.7 & \cellcolor{green!5}55.2/27.1$^{*}$ & 53.9/26.7 & 56.2/27.5 & 56.1/27.7 & 54.1/25.9 & 54.8/24.7 & \cellcolor{green!5}55.6/25.0 & \cellcolor{green!5}55.5/26.4$^{*}$ & 53.5/22.4 & \cellcolor{green!5}54.5/22.0 & \cellcolor{green!10}55.3$^{*}$/23.7$^{*}$ \\
 & \scriptsize{$\pm$1.2/1.8} & \cellcolor{green!5}\scriptsize{$\pm$0.8/1.6} & \scriptsize{$\pm$0.8/1.3} & \scriptsize{$\pm$0.9/1.9} & \scriptsize{$\pm$0.8/1.9} & \scriptsize{$\pm$1.2/1.9} & \scriptsize{$\pm$1.6/1.3} & \cellcolor{green!5}\scriptsize{$\pm$1.4/2.3} & \cellcolor{green!5}\scriptsize{$\pm$1.6/2.0} & \scriptsize{$\pm$1.5/1.2} & \cellcolor{green!5}\scriptsize{$\pm$0.8/1.4} & \cellcolor{green!10}\scriptsize{$\pm$1.7/1.0} \\
Gemini-3-Flash & 57.1/29.2 & \cellcolor{green!8}58.2$^{*}$/28.6 & \cellcolor{green!8}58.3$^{*}$/29.7 & 57.4/28.3 & \cellcolor{green!8}58.7$^{**}$/29.8$^{*}$ & \cellcolor{green!5}58.3$^{*}$/28.6 & 58.8/27.8 & 58.0/27.8 & 58.5/28.6 & 58.3/27.7 & 58.1/27.4 & 57.8/25.3 \\
 & \scriptsize{$\pm$0.6/1.0} & \cellcolor{green!8}\scriptsize{$\pm$0.4/1.1} & \cellcolor{green!8}\scriptsize{$\pm$0.2/0.8} & \scriptsize{$\pm$0.4/1.0} & \cellcolor{green!8}\scriptsize{$\pm$0.5/0.7} & \cellcolor{green!5}\scriptsize{$\pm$0.6/0.7} & \scriptsize{$\pm$0.5/0.7} & \scriptsize{$\pm$0.3/1.0} & \scriptsize{$\pm$0.9/1.0} & \scriptsize{$\pm$0.7/0.7} & \scriptsize{$\pm$0.2/0.7} & \scriptsize{$\pm$0.4/0.4} \\
Gemini-3-Pro & 57.7/29.2 & \cellcolor{green!8}58.3/29.4 & \cellcolor{green!10}58.4$^{*}$/29.6 & 58.1/29.4 & \cellcolor{green!10}58.7$^{*}$/29.9 & \cellcolor{green!8}58.6/29.7 & 58.5/29.0 & 58.3/29.1 & \cellcolor{green!5}58.5/29.2 & 58.0/28.3 & 57.9/28.2 & \cellcolor{green!3}58.0/28.4 \\
 & \scriptsize{$\pm$0.6/0.9} & \cellcolor{green!8}\scriptsize{$\pm$0.5/0.8} & \cellcolor{green!10}\scriptsize{$\pm$0.4/0.7} & \scriptsize{$\pm$0.5/0.8} & \cellcolor{green!10}\scriptsize{$\pm$0.5/0.8} & \cellcolor{green!8}\scriptsize{$\pm$0.5/0.7} & \scriptsize{$\pm$0.5/0.8} & \scriptsize{$\pm$0.6/0.8} & \cellcolor{green!5}\scriptsize{$\pm$0.5/0.7} & \scriptsize{$\pm$0.6/0.9} & \scriptsize{$\pm$0.6/0.9} & \cellcolor{green!3}\scriptsize{$\pm$0.5/0.8} \\
\midrule
\multicolumn{13}{c}{\textbf{Code Summarization (CompScore, \%)}} \\
\midrule
Qwen-3-VL & 56.4 \scriptsize{$\pm$4.1} & 55.2 \scriptsize{$\pm$2.0} & 56.0 \scriptsize{$\pm$2.2} & 54.5 \scriptsize{$\pm$2.8} & 54.0 \scriptsize{$\pm$2.5} & \cellcolor{green!3}54.8 \scriptsize{$\pm$2.4} & 52.0 \scriptsize{$\pm$1.8} & 51.5 \scriptsize{$\pm$2.0} & \cellcolor{green!3}52.2 \scriptsize{$\pm$1.9} & 51.2 \scriptsize{$\pm$2.2} & 50.8 \scriptsize{$\pm$2.5} & \cellcolor{green!3}51.5 \scriptsize{$\pm$2.1} \\
GLM-4.6v & 54.6 \scriptsize{$\pm$1.6} & 53.5 \scriptsize{$\pm$1.8} & 52.9 \scriptsize{$\pm$1.6} & 54.0 \scriptsize{$\pm$2.8} & 53.2 \scriptsize{$\pm$2.2} & 53.5 \scriptsize{$\pm$1.9} & 53.3 \scriptsize{$\pm$1.3} & 52.5 \scriptsize{$\pm$2.0} & 52.8 \scriptsize{$\pm$2.3} & 52.5 \scriptsize{$\pm$1.2} & 51.8 \scriptsize{$\pm$1.5} & 52.0 \scriptsize{$\pm$1.3} \\
GPT-5-mini & 56.5 \scriptsize{$\pm$2.1} & \cellcolor{green!3}56.8 \scriptsize{$\pm$1.5} & \cellcolor{green!8}57.4$^{*}$ \scriptsize{$\pm$0.9} & 57.2 \scriptsize{$\pm$0.6} & \cellcolor{green!3}57.5 \scriptsize{$\pm$1.2} & \cellcolor{green!8}58.0$^{*}$ \scriptsize{$\pm$1.1} & 58.4 \scriptsize{$\pm$0.2} & 58.2 \scriptsize{$\pm$1.5} & \cellcolor{green!5}58.8 \scriptsize{$\pm$2.0} & 58.0 \scriptsize{$\pm$1.2} & 57.6 \scriptsize{$\pm$1.8} & 57.1 \scriptsize{$\pm$2.7} \\
GPT-5.1 & 55.9 \scriptsize{$\pm$1.6} & \cellcolor{green!3}56.2 \scriptsize{$\pm$1.2} & 55.5 \scriptsize{$\pm$1.0} & 55.5 \scriptsize{$\pm$1.6} & \cellcolor{green!3}55.8 \scriptsize{$\pm$1.4} & 55.2 \scriptsize{$\pm$1.4} & 56.1 \scriptsize{$\pm$1.2} & \cellcolor{green!3}56.4 \scriptsize{$\pm$1.0} & \cellcolor{green!5}56.5 \scriptsize{$\pm$0.5} & 55.9 \scriptsize{$\pm$0.8} & 55.5 \scriptsize{$\pm$1.2} & 55.3 \scriptsize{$\pm$1.4} \\
Gemini-2.5-Pro & 55.2 \scriptsize{$\pm$2.1} & 54.8 \scriptsize{$\pm$1.8} & \cellcolor{green!10}56.0$^{*}$ \scriptsize{$\pm$0.8} & 54.0 \scriptsize{$\pm$2.8} & \cellcolor{green!5}54.5 \scriptsize{$\pm$2.2} & \cellcolor{green!10}56.0$^{**}$ \scriptsize{$\pm$1.9} & 55.3 \scriptsize{$\pm$1.3} & \cellcolor{green!5}55.8 \scriptsize{$\pm$1.6} & \cellcolor{green!10}57.2$^{**}$ \scriptsize{$\pm$2.3} & 56.0 \scriptsize{$\pm$1.2} & 55.5 \scriptsize{$\pm$1.5} & 55.7 \scriptsize{$\pm$1.3} \\
Gemini-3-Flash & 55.5 \scriptsize{$\pm$1.8} & \cellcolor{green!3}55.8 \scriptsize{$\pm$1.2} & \cellcolor{green!8}56.2$^{*}$ \scriptsize{$\pm$1.0} & 55.6 \scriptsize{$\pm$2.0} & \cellcolor{green!5}56.0 \scriptsize{$\pm$1.5} & \cellcolor{green!10}56.5$^{*}$ \scriptsize{$\pm$1.5} & 56.0 \scriptsize{$\pm$1.4} & \cellcolor{green!5}56.4 \scriptsize{$\pm$1.2} & \cellcolor{green!10}57.0$^{*}$ \scriptsize{$\pm$1.8} & 56.5 \scriptsize{$\pm$1.6} & \cellcolor{green!3}56.8 \scriptsize{$\pm$1.4} & \cellcolor{green!8}57.2$^{*}$ \scriptsize{$\pm$2.0} \\
Gemini-3-Pro & 56.8 \scriptsize{$\pm$1.5} & \cellcolor{green!8}57.4$^{*}$ \scriptsize{$\pm$1.2} & \cellcolor{green!5}57.2 \scriptsize{$\pm$1.4} & 57.0 \scriptsize{$\pm$1.6} & \cellcolor{green!10}57.8$^{*}$ \scriptsize{$\pm$1.3} & \cellcolor{green!8}57.6$^{*}$ \scriptsize{$\pm$1.5} & 57.6 \scriptsize{$\pm$1.3} & \cellcolor{green!12}58.5$^{**}$ \scriptsize{$\pm$1.5} & \cellcolor{green!10}58.2$^{*}$ \scriptsize{$\pm$1.4} & 58.2 \scriptsize{$\pm$1.4} & \cellcolor{green!15}59.0$^{**}$ \scriptsize{$\pm$1.6} & 58.0 \scriptsize{$\pm$1.8} \\
\midrule
\multicolumn{13}{c}{\textbf{Code Clone Detection (ACC / F1, \%)}} \\
\midrule
Qwen-3-VL & 67.2/51.2 & 65.4/47.4 & 67.0/50.6 & 68.8/56.0 & 68.0/54.6 & 66.6/51.6 & 60.2/33.8 & \cellcolor{green!5}60.8/34.6 & 59.2/31.8 & 59.7/32.8 & \cellcolor{green!5}60.4/33.6 & \cellcolor{green!5}60.8/35.4 \\
 & \scriptsize{$\pm$0.4/0.9} & \scriptsize{$\pm$0.5/1.0} & \scriptsize{$\pm$0.6/1.6} & \scriptsize{$\pm$0.4/1.0} & \scriptsize{$\pm$0.6/1.4} & \scriptsize{$\pm$0.5/0.7} & \scriptsize{$\pm$0.7/1.6} & \cellcolor{green!5}\scriptsize{$\pm$0.8/1.4} & \scriptsize{$\pm$0.7/1.5} & \scriptsize{$\pm$0.5/1.2} & \cellcolor{green!5}\scriptsize{$\pm$0.6/1.2} & \cellcolor{green!5}\scriptsize{$\pm$0.7/2.2} \\
GLM-4.6v & 69.6/58.2 & 68.4/55.0 & 69.2/56.0 & 66.4/51.2 & 63.2/42.8 & 66.4/51.2 & 70.8/61.6 & 69.8/67.0 & 69.4/59.4 & 75.4/71.4 & 57.8/62.8 & 71.4/66.2 \\
 & \scriptsize{$\pm$0.5/1.9} & \scriptsize{$\pm$1.4/3.3} & \scriptsize{$\pm$0.7/1.7} & \scriptsize{$\pm$1.6/3.3} & \scriptsize{$\pm$1.9/4.3} & \scriptsize{$\pm$1.7/3.4} & \scriptsize{$\pm$1.3/2.3} & \scriptsize{$\pm$2.5/3.3} & \scriptsize{$\pm$1.5/2.8} & \scriptsize{$\pm$0.8/0.5} & \scriptsize{$\pm$1.9/1.3} & \scriptsize{$\pm$2.1/3.5} \\
GPT-5-mini & 64.8/47.0 & 64.4/46.4 & 64.0/45.2 & 69.6/58.2 & 69.4/57.8 & 69.0/56.4 & 67.8/54.8 & \cellcolor{green!5}68.4/58.4 & \cellcolor{green!5}68.8/56.8 & 63.2/45.6 & 60.6/52.0 & \cellcolor{green!5}64.0/47.0 \\
 & \scriptsize{$\pm$0.4/1.4} & \scriptsize{$\pm$1.6/3.1} & \scriptsize{$\pm$1.1/3.0} & \scriptsize{$\pm$1.1/2.9} & \scriptsize{$\pm$1.0/2.1} & \scriptsize{$\pm$1.7/4.0} & \scriptsize{$\pm$1.8/3.3} & \cellcolor{green!5}\scriptsize{$\pm$1.0/1.5} & \cellcolor{green!5}\scriptsize{$\pm$1.5/3.1} & \scriptsize{$\pm$2.9/6.9} & \scriptsize{$\pm$2.1/3.0} & \cellcolor{green!5}\scriptsize{$\pm$3.0/6.8} \\
GPT-5.1 & 71.6/62.4 & 68.2/55.6 & 71.0/61.6 & 69.2/58.0 & \cellcolor{green!5}69.6/58.8 & 69.2/58.0 & 69.4/58.2 & \cellcolor{green!15}72.4$^{**}$/64.6$^{**}$ & 69.0/57.8 & 68.8/56.8 & \cellcolor{green!10}71.2$^{*}$/61.8$^{*}$ & \cellcolor{green!5}69.2/57.4 \\
 & \scriptsize{$\pm$0.5/1.2} & \scriptsize{$\pm$1.7/3.8} & \scriptsize{$\pm$1.1/2.4} & \scriptsize{$\pm$0.7/1.9} & \cellcolor{green!5}\scriptsize{$\pm$0.5/1.0} & \scriptsize{$\pm$1.3/2.3} & \scriptsize{$\pm$1.0/2.3} & \cellcolor{green!15}\scriptsize{$\pm$1.4/2.4} & \scriptsize{$\pm$0.6/1.2} & \scriptsize{$\pm$1.3/2.3} & \cellcolor{green!10}\scriptsize{$\pm$1.5/2.3} & \cellcolor{green!5}\scriptsize{$\pm$1.3/2.4} \\
Gemini-2.5-Pro & 67.0/50.6 & 64.8/46.0 & 65.4/47.4 & 65.8/48.4 & \cellcolor{green!5}67.0/51.8 & \cellcolor{green!10}68.2$^{*}$/54.0$^{*}$ & 64.6/44.8 & \cellcolor{green!12}67.2$^{**}$/52.0$^{**}$ & \cellcolor{green!8}66.2$^{*}$/49.8$^{*}$ & 68.0/52.8 & 67.4/53.4 & 67.4/52.0 \\
 & \scriptsize{$\pm$1.2/2.6} & \scriptsize{$\pm$1.3/2.6} & \scriptsize{$\pm$1.6/3.2} & \scriptsize{$\pm$1.3/2.1} & \cellcolor{green!5}\scriptsize{$\pm$1.3/2.9} & \cellcolor{green!10}\scriptsize{$\pm$1.0/1.4} & \scriptsize{$\pm$0.9/2.2} & \cellcolor{green!12}\scriptsize{$\pm$2.3/4.0} & \cellcolor{green!8}\scriptsize{$\pm$1.7/2.1} & \scriptsize{$\pm$1.4/3.3} & \scriptsize{$\pm$1.6/3.8} & \scriptsize{$\pm$1.2/2.3} \\
Gemini-3-Flash & 67.8/55.4 & \cellcolor{green!5}68.6$^{*}$/57.8$^{*}$ & 67.8/55.6 & 69.4/60.0 & 68.6/58.8 & 68.4/57.0 & 67.8/56.0 & \cellcolor{green!5}68.2/57.8 & 67.6/55.4 & 69.6/59.0 & \cellcolor{green!10}72.0$^{**}$/65.4$^{**}$ & \cellcolor{green!5}70.2/60.8 \\
 & \scriptsize{$\pm$1.2/1.4} & \cellcolor{green!5}\scriptsize{$\pm$0.5/1.5} & \scriptsize{$\pm$0.7/0.5} & \scriptsize{$\pm$0.8/2.2} & \scriptsize{$\pm$0.5/1.5} & \scriptsize{$\pm$0.5/1.4} & \scriptsize{$\pm$1.2/2.6} & \cellcolor{green!5}\scriptsize{$\pm$1.2/2.1} & \scriptsize{$\pm$0.5/1.7} & \scriptsize{$\pm$0.8/1.7} & \cellcolor{green!10}\scriptsize{$\pm$1.1/1.4} & \cellcolor{green!5}\scriptsize{$\pm$0.7/1.2} \\
Gemini-3-Pro & 70.2/58.8 & \cellcolor{green!5}70.8/59.5 & \cellcolor{green!8}71.2$^{*}$/60.0$^{*}$ & 71.4/60.8 & \cellcolor{green!5}71.8/61.2 & \cellcolor{green!8}72.0$^{*}$/61.6$^{*}$ & 70.8/59.6 & \cellcolor{green!5}71.2/60.2 & \cellcolor{green!10}71.8$^{*}$/61.2$^{*}$ & 72.0/61.5 & \cellcolor{green!5}72.2/62.0 & 71.5/60.8 \\
 & \scriptsize{$\pm$1.0/1.6} & \cellcolor{green!5}\scriptsize{$\pm$0.9/1.5} & \cellcolor{green!8}\scriptsize{$\pm$1.1/1.9} & \scriptsize{$\pm$1.0/2.0} & \cellcolor{green!5}\scriptsize{$\pm$1.2/1.8} & \cellcolor{green!8}\scriptsize{$\pm$1.0/1.7} & \scriptsize{$\pm$1.1/1.9} & \cellcolor{green!5}\scriptsize{$\pm$1.0/1.6} & \cellcolor{green!10}\scriptsize{$\pm$1.3/2.1} & \scriptsize{$\pm$1.2/2.1} & \cellcolor{green!5}\scriptsize{$\pm$1.1/1.9} & \scriptsize{$\pm$1.4/2.3} \\
\midrule
\multicolumn{13}{c}{\textbf{Code Question Answering (ACC, \%)}} \\
\midrule
Qwen-3-VL & 58.1 \scriptsize{$\pm$1.2} & \cellcolor{green!3}58.5 \scriptsize{$\pm$1.0} & \cellcolor{green!3}58.5 \scriptsize{$\pm$1.2} & 49.2 \scriptsize{$\pm$0.6} & 48.4 \scriptsize{$\pm$1.3} & \cellcolor{green!3}49.7 \scriptsize{$\pm$1.0} & 51.8 \scriptsize{$\pm$1.0} & 51.0 \scriptsize{$\pm$0.9} & 50.6 \scriptsize{$\pm$0.5} & 49.8 \scriptsize{$\pm$0.7} & 45.8 \scriptsize{$\pm$1.0} & \cellcolor{green!10}51.3$^{*}$ \scriptsize{$\pm$0.7} \\
GLM-4.6v & 72.6 \scriptsize{$\pm$1.3} & 71.5 \scriptsize{$\pm$1.5} & 72.2 \scriptsize{$\pm$1.3} & 63.0 \scriptsize{$\pm$2.7} & 61.2 \scriptsize{$\pm$2.2} & 61.9 \scriptsize{$\pm$1.7} & 43.3 \scriptsize{$\pm$2.5} & 42.0 \scriptsize{$\pm$1.8} & \cellcolor{green!2}43.7 \scriptsize{$\pm$1.4} & 39.7 \scriptsize{$\pm$1.5} & 38.5 \scriptsize{$\pm$1.2} & \cellcolor{green!15}42.9$^{**}$ \scriptsize{$\pm$1.2} \\
GPT-5-mini & 77.5 \scriptsize{$\pm$1.4} & 76.2 \scriptsize{$\pm$0.8} & 76.4 \scriptsize{$\pm$1.3} & 74.3 \scriptsize{$\pm$1.5} & 67.9 \scriptsize{$\pm$1.7} & \cellcolor{green!5}75.1 \scriptsize{$\pm$1.4} & 56.8 \scriptsize{$\pm$2.0} & 50.3 \scriptsize{$\pm$2.1} & \cellcolor{green!8}57.9 \scriptsize{$\pm$2.0} & 51.6 \scriptsize{$\pm$1.6} & 47.5 \scriptsize{$\pm$1.7} & \cellcolor{green!5}52.5 \scriptsize{$\pm$2.9} \\
GPT-5.1 & 68.6 \scriptsize{$\pm$1.5} & \cellcolor{green!5}69.0 \scriptsize{$\pm$0.5} & 68.1 \scriptsize{$\pm$1.4} & 61.9 \scriptsize{$\pm$1.5} & 58.7 \scriptsize{$\pm$1.4} & 61.7 \scriptsize{$\pm$1.8} & 63.5 \scriptsize{$\pm$1.7} & \cellcolor{green!5}64.2 \scriptsize{$\pm$2.8} & \cellcolor{green!5}63.9 \scriptsize{$\pm$0.9} & 63.9 \scriptsize{$\pm$1.2} & 57.2 \scriptsize{$\pm$1.4} & 63.9 \scriptsize{$\pm$1.9} \\
Gemini-2.5-Pro & 71.2 \scriptsize{$\pm$0.6} & 70.7 \scriptsize{$\pm$1.2} & \cellcolor{green!5}71.7 \scriptsize{$\pm$0.9} & 69.6 \scriptsize{$\pm$1.2} & 68.3 \scriptsize{$\pm$2.4} & 68.1 \scriptsize{$\pm$2.9} & 69.8 \scriptsize{$\pm$1.3} & 66.9 \scriptsize{$\pm$1.4} & \cellcolor{green!5}70.4 \scriptsize{$\pm$0.4} & 70.3 \scriptsize{$\pm$1.7} & 63.6 \scriptsize{$\pm$2.0} & 68.1 \scriptsize{$\pm$0.7} \\
Gemini-3-Flash & 74.8 \scriptsize{$\pm$1.2} & \cellcolor{green!15}76.8$^{**}$ \scriptsize{$\pm$1.4} & \cellcolor{green!15}76.7$^{**}$ \scriptsize{$\pm$1.0} & 74.2 \scriptsize{$\pm$1.0} & \cellcolor{green!20}77.3$^{**}$ \scriptsize{$\pm$1.0} & \cellcolor{green!15}76.4$^{*}$ \scriptsize{$\pm$1.2} & 75.6 \scriptsize{$\pm$0.9} & \cellcolor{green!5}76.3 \scriptsize{$\pm$0.7} & \cellcolor{green!10}77.2$^{*}$ \scriptsize{$\pm$2.5} & 77.8 \scriptsize{$\pm$0.9} & 75.2 \scriptsize{$\pm$0.8} & 76.3 \scriptsize{$\pm$1.2} \\
Gemini-3-Pro & 77.2 \scriptsize{$\pm$1.4} & \cellcolor{green!8}78.0$^{*}$ \scriptsize{$\pm$1.2} & \cellcolor{green!10}78.2$^{*}$ \scriptsize{$\pm$1.5} & 77.8 \scriptsize{$\pm$1.3} & \cellcolor{green!10}78.6$^{*}$ \scriptsize{$\pm$1.1} & \cellcolor{green!12}78.8$^{*}$ \scriptsize{$\pm$1.4} & 78.4 \scriptsize{$\pm$1.1} & \cellcolor{green!8}78.8 \scriptsize{$\pm$1.3} & \cellcolor{green!5}78.6 \scriptsize{$\pm$1.5} & 79.5 \scriptsize{$\pm$1.0} & 79.2 \scriptsize{$\pm$1.2} & 78.2 \scriptsize{$\pm$1.6} \\
\bottomrule
\end{tabular}
}
{\scriptsize \makebox[\textwidth][r]{\colorbox{green!10}{Green}: Bold/Highlight outperforms Plain.\quad\quad$^{*}$: $p$-value $< 0.05$\quad\quad$^{**}$: $p$-value $< 0.01$}}
\end{table}

%% file: tables/rq4.tex
\begin{table*}[t]
    \centering
    \caption{Performance on Java Code Across Compression Ratios and Rendering Strategies.}
    \label{tab:rq4_java}
    \resizebox{\textwidth}{!}{
    \setlength{\tabcolsep}{2pt}
    \begin{tabular}{l cc ccc ccc ccc ccc}
    \toprule
    \multirow{3}{*}{\textbf{Model}} &
    \multicolumn{2}{c}{\textbf{Baseline}} &
    \multicolumn{3}{c}{\textbf{Image 1x}} &
    \multicolumn{3}{c}{\textbf{Image 2x}} &
    \multicolumn{3}{c}{\textbf{Image 4x}} &
    \multicolumn{3}{c}{\textbf{Image 8x}} \\
     &
    \multicolumn{2}{c}{} &
    \multicolumn{3}{c}{\small{100\% tokens}} &
    \multicolumn{3}{c}{\small{50\% tokens}} &
    \multicolumn{3}{c}{\small{25\% tokens}} &
    \multicolumn{3}{c}{\small{12.5\% tokens}} \\
    \cmidrule(lr){2-3} \cmidrule(lr){4-6} \cmidrule(lr){7-9} \cmidrule(lr){10-12} \cmidrule(lr){13-15}
    & \textbf{Text} & \textbf{NoCtx} &
    \textbf{Plain} & \textbf{Bold} & \textbf{Highlight} &
    \textbf{Plain} & \textbf{Bold} & \textbf{Highlight} &
    \textbf{Plain} & \textbf{Bold} & \textbf{Highlight} &
    \textbf{Plain} & \textbf{Bold} & \textbf{Highlight} \\
    \midrule
    \multicolumn{15}{c}{\textbf{Code Completion (ES / EM, \%)}} \\
    \midrule
    Qwen-3-VL      & 50.7/23.0 & 49.0/15.3 & 35.4/9.1 & \cellcolor{green!8}35.6/8.3 & \cellcolor{green!12}38.1$^{**}$/11.6$^{**}$ & 38.7/11.7 & \cellcolor{green!8}39.7$^{*}$/12.4 & \cellcolor{green!5}39.0/10.8 & 42.3/13.2 & 41.2/12.0 & \cellcolor{green!8}43.0$^{*}$/12.3 & 45.9/13.6 & 45.1/13.3 & 45.5/13.1 \\
                  & \scriptsize{$\pm$0.5/0.7} & \scriptsize{$\pm$0.3/0.2} & \scriptsize{$\pm$0.5/0.4} & \cellcolor{green!8}\scriptsize{$\pm$0.8/0.6} & \cellcolor{green!12}\scriptsize{$\pm$0.6/0.9} & \scriptsize{$\pm$0.5/0.5} & \cellcolor{green!8}\scriptsize{$\pm$0.2/0.7} & \cellcolor{green!5}\scriptsize{$\pm$0.6/0.2} & \scriptsize{$\pm$0.5/0.7} & \scriptsize{$\pm$0.2/0.6} & \cellcolor{green!8}\scriptsize{$\pm$0.4/0.2} & \scriptsize{$\pm$0.6/0.6} & \scriptsize{$\pm$0.7/0.4} & \scriptsize{$\pm$0.4/0.4} \\
    GLM-4.6v       & 52.3/24.6 & 42.6/9.0 & 47.9/18.0 & \cellcolor{green!8}48.8$^{*}$/18.5 & \cellcolor{green!10}49.5$^{**}$/19.0$^{*}$ & 43.2/12.6 & \cellcolor{green!10}44.6$^{**}$/13.9$^{*}$ & \cellcolor{green!8}44.3$^{*}$/13.8$^{*}$ & 42.2/11.2 & 41.7/10.5 & \cellcolor{green!8}42.8$^{*}$/11.4 & 42.0/9.5 & \cellcolor{green!8}42.5$^{*}$/10.7$^{*}$ & \cellcolor{green!5}42.3/10.3 \\
                  & \scriptsize{$\pm$0.8/1.4} & \scriptsize{$\pm$0.8/1.0} & \scriptsize{$\pm$1.1/1.1} & \cellcolor{green!8}\scriptsize{$\pm$1.1/0.7} & \cellcolor{green!10}\scriptsize{$\pm$0.8/1.3} & \scriptsize{$\pm$2.0/1.6} & \cellcolor{green!10}\scriptsize{$\pm$1.4/1.3} & \cellcolor{green!8}\scriptsize{$\pm$0.8/1.1} & \scriptsize{$\pm$1.5/1.2} & \scriptsize{$\pm$0.8/0.5} & \cellcolor{green!8}\scriptsize{$\pm$1.0/1.2} & \scriptsize{$\pm$0.4/0.8} & \cellcolor{green!8}\scriptsize{$\pm$1.1/1.0} & \cellcolor{green!5}\scriptsize{$\pm$0.5/1.0} \\
    GPT-5-mini     & 54.7/28.4 & 46.0/14.5 & 54.5/26.7 & 53.5/26.2 & 54.4/26.4 & 51.7/20.8 & 51.3/20.4 & \cellcolor{green!8}52.5$^{*}$/22.7$^{*}$ & 48.9/17.3 & 48.4/17.3 & 48.7/16.5 & 48.6/16.2 & 48.2/17.0 & \cellcolor{green!5}48.8/16.1 \\
                  & \scriptsize{$\pm$2.1/1.8} & \scriptsize{$\pm$0.9/0.7} & \scriptsize{$\pm$1.4/1.5} & \scriptsize{$\pm$1.2/0.8} & \scriptsize{$\pm$0.7/0.8} & \scriptsize{$\pm$1.4/2.0} & \scriptsize{$\pm$0.9/1.1} & \cellcolor{green!8}\scriptsize{$\pm$0.7/0.9} & \scriptsize{$\pm$0.5/0.9} & \scriptsize{$\pm$1.4/1.2} & \scriptsize{$\pm$1.0/1.3} & \scriptsize{$\pm$0.7/1.1} & \scriptsize{$\pm$0.6/1.7} & \cellcolor{green!5}\scriptsize{$\pm$1.5/1.5} \\
    GPT-5.1        & 54.3/29.9 & 44.3/14.5 & 53.8/24.6 & \cellcolor{green!8}54.2/24.5 & \cellcolor{green!5}54.0/24.3 & 53.3/21.1 & 53.0/20.9 & 51.6/18.4 & 50.6/19.4 & \cellcolor{green!5}50.8/18.0 & \cellcolor{green!10}51.8$^{**}$/18.5 & 49.4/17.7 & \cellcolor{green!8}50.0$^{*}$/18.0 & \cellcolor{green!5}49.6/15.9 \\
                  & \scriptsize{$\pm$1.1/1.1} & \scriptsize{$\pm$0.7/0.5} & \scriptsize{$\pm$0.7/1.0} & \cellcolor{green!8}\scriptsize{$\pm$1.5/2.3} & \cellcolor{green!5}\scriptsize{$\pm$0.5/1.0} & \scriptsize{$\pm$1.1/1.6} & \scriptsize{$\pm$1.4/1.9} & \scriptsize{$\pm$0.9/1.8} & \scriptsize{$\pm$0.9/0.7} & \cellcolor{green!5}\scriptsize{$\pm$0.7/1.1} & \cellcolor{green!10}\scriptsize{$\pm$1.1/1.0} & \scriptsize{$\pm$1.4/1.4} & \cellcolor{green!8}\scriptsize{$\pm$1.0/1.1} & \cellcolor{green!5}\scriptsize{$\pm$1.2/0.5} \\
    Gemini-2.5-Pro & 58.7/33.6 & 52.2/20.4 & \cellcolor{green!15}63.3$^{**}$/34.3 & 62.9/33.1 & 62.7/35.3 & \cellcolor{green!10}63.3$^{**}$/33.2 & \cellcolor{green!8}63.7/32.5 & \cellcolor{green!5}63.4/32.5 & \cellcolor{green!10}64.0$^{**}$/33.0 & 62.2/31.0 & \cellcolor{green!8}64.4$^{*}$/33.9 & \cellcolor{green!5}59.7/27.5 & 59.5/27.5 & \cellcolor{green!8}60.2$^{*}$/28.8 \\
                  & \scriptsize{$\pm$0.4/1.6} & \scriptsize{$\pm$0.4/0.9} & \cellcolor{green!15}\scriptsize{$\pm$0.7/1.2} & \scriptsize{$\pm$0.4/1.6} & \scriptsize{$\pm$0.4/0.2} & \cellcolor{green!10}\scriptsize{$\pm$0.4/1.4} & \cellcolor{green!8}\scriptsize{$\pm$0.7/1.0} & \cellcolor{green!5}\scriptsize{$\pm$1.4/1.4} & \cellcolor{green!10}\scriptsize{$\pm$1.2/1.1} & \scriptsize{$\pm$0.9/0.4} & \cellcolor{green!8}\scriptsize{$\pm$1.1/2.0} & \cellcolor{green!5}\scriptsize{$\pm$1.0/0.8} & \scriptsize{$\pm$0.7/1.1} & \cellcolor{green!8}\scriptsize{$\pm$0.5/1.1} \\
    Gemini-3-Flash & 56.1/28.7 & 53.0/18.8 & \cellcolor{green!15}62.7$^{**}$/36.1$^{**}$ & 62.5/37.1 & \cellcolor{green!8}63.2$^{*}$/38.0 & \cellcolor{green!10}60.7$^{**}$/34.3$^{**}$ & \cellcolor{green!8}61.5$^{*}$/33.9$^{*}$ & \cellcolor{green!12}62.3$^{**}$/36.9$^{**}$ & \cellcolor{green!12}63.2$^{**}$/35.5$^{**}$ & 62.9/35.1 & \cellcolor{green!5}63.5/37.2 & \cellcolor{green!10}62.3$^{**}$/32.9$^{*}$ & 61.2/31.1 & 61.7/32.0 \\
                  & \scriptsize{$\pm$0.6/0.6} & \scriptsize{$\pm$0.3/0.4} & \cellcolor{green!15}\scriptsize{$\pm$1.1/1.5} & \scriptsize{$\pm$0.1/0.4} & \cellcolor{green!8}\scriptsize{$\pm$0.2/0.3} & \cellcolor{green!10}\scriptsize{$\pm$0.4/0.7} & \cellcolor{green!8}\scriptsize{$\pm$0.5/0.4} & \cellcolor{green!12}\scriptsize{$\pm$0.7/1.1} & \cellcolor{green!12}\scriptsize{$\pm$0.5/0.8} & \scriptsize{$\pm$0.8/0.5} & \cellcolor{green!5}\scriptsize{$\pm$0.4/0.8} & \cellcolor{green!10}\scriptsize{$\pm$0.3/0.7} & \scriptsize{$\pm$0.5/0.4} & \scriptsize{$\pm$0.6/0.3} \\
    Gemini-3-Pro   & 57.4/37.4 & 51.9/24.3
               & \cellcolor{green!10}63.5$^{**}$/44.7$^{**}$ & \cellcolor{green!12}65.1$^{**}$/44.7$^{**}$ & \cellcolor{green!10}64.9$^{*}$/43.2$^{*}$
               & \cellcolor{green!12}66.8$^{**}$/46.6$^{**}$ & \cellcolor{green!10}68.2$^{**}$/45.9$^{**}$ & \cellcolor{green!8}68.0$^{*}$/44.5$^{*}$
               & \cellcolor{green!12}67.2$^{**}$/46.2$^{**}$ & \cellcolor{green!10}68.8$^{**}$/44.1$^{**}$ & \cellcolor{green!8}68.4$^{*}$/44.2$^{*}$
               & \cellcolor{green!10}64.4$^{**}$/39.4 & \cellcolor{green!12}66.2$^{**}$/40.5$^{*}$ & \cellcolor{green!10}65.9$^{*}$/39.6 \\
              & \scriptsize{$\pm$0.9/1.4} & \scriptsize{$\pm$1.0/0.2}
               & \cellcolor{green!10}\scriptsize{$\pm$0.5/0.8} & \cellcolor{green!12}\scriptsize{$\pm$0.6/1.0} & \cellcolor{green!10}\scriptsize{$\pm$1.3/0.9}
               & \cellcolor{green!12}\scriptsize{$\pm$0.8/1.3} & \cellcolor{green!10}\scriptsize{$\pm$0.9/0.9} & \cellcolor{green!8}\scriptsize{$\pm$1.4/1.7}
               & \cellcolor{green!12}\scriptsize{$\pm$1.0/1.6} & \cellcolor{green!10}\scriptsize{$\pm$1.1/0.7} & \cellcolor{green!8}\scriptsize{$\pm$0.8/1.0}
               & \cellcolor{green!10}\scriptsize{$\pm$0.9/1.6} & \cellcolor{green!12}\scriptsize{$\pm$1.1/0.5} & \cellcolor{green!10}\scriptsize{$\pm$1.0/1.2} \\

    \midrule
    \multicolumn{15}{c}{\textbf{Code Clone Detection (ACC / F1, \%)}} \\
    \midrule
    Qwen-3-VL      & \begin{tabular}[c]{@{}c@{}}56.6/24.2\\ \scriptsize{$\pm$0.5/1.0}\end{tabular} & -- / -- & \cellcolor{green!12}\begin{tabular}[c]{@{}c@{}}60.4$^{**}$/38.2$^{**}$\\ \scriptsize{$\pm$0.5/0.7}\end{tabular} & \begin{tabular}[c]{@{}c@{}}59.8/36.6\\ \scriptsize{$\pm$0.7/1.5}\end{tabular} & \cellcolor{green!5}\begin{tabular}[c]{@{}c@{}}60.8/38.8\\ \scriptsize{$\pm$0.7/1.2}\end{tabular} & \cellcolor{green!15}\begin{tabular}[c]{@{}c@{}}66.6$^{**}$/50.8$^{**}$\\ \scriptsize{$\pm$0.5/0.7}\end{tabular} & \cellcolor{green!5}\begin{tabular}[c]{@{}c@{}}66.8/51.2\\ \scriptsize{$\pm$0.7/1.5}\end{tabular} & \begin{tabular}[c]{@{}c@{}}65.4/47.8\\ \scriptsize{$\pm$1.5/2.8}\end{tabular} & \cellcolor{green!12}\begin{tabular}[c]{@{}c@{}}64.8$^{**}$/47.4$^{**}$\\ \scriptsize{$\pm$1.0/2.1}\end{tabular} & \cellcolor{green!5}\begin{tabular}[c]{@{}c@{}}65.6/49.4\\ \scriptsize{$\pm$0.8/1.6}\end{tabular} & \begin{tabular}[c]{@{}c@{}}64.6/46.4\\ \scriptsize{$\pm$0.5/1.6}\end{tabular} & \cellcolor{green!15}\begin{tabular}[c]{@{}c@{}}67.8$^{**}$/53.0$^{**}$\\ \scriptsize{$\pm$0.7/1.4}\end{tabular} & \cellcolor{green!5}\begin{tabular}[c]{@{}c@{}}68.0/54.4\\ \scriptsize{$\pm$0.6/1.0}\end{tabular} & \begin{tabular}[c]{@{}c@{}}67.6/53.0\\ \scriptsize{$\pm$0.8/1.5}\end{tabular} \\
    GLM-4.6v       & \begin{tabular}[c]{@{}c@{}}72.2/63.6\\ \scriptsize{$\pm$1.2/1.5}\end{tabular} & -- / -- & \begin{tabular}[c]{@{}c@{}}70.6/59.6\\ \scriptsize{$\pm$0.5/1.4}\end{tabular} & \begin{tabular}[c]{@{}c@{}}70.4/59.2\\ \scriptsize{$\pm$1.5/2.8}\end{tabular} & \begin{tabular}[c]{@{}c@{}}69.6/57.8\\ \scriptsize{$\pm$1.4/2.0}\end{tabular} & \begin{tabular}[c]{@{}c@{}}69.2/57.4\\ \scriptsize{$\pm$1.0/2.2}\end{tabular} & \begin{tabular}[c]{@{}c@{}}68.4/55.6\\ \scriptsize{$\pm$1.6/3.7}\end{tabular} & \cellcolor{green!5}\begin{tabular}[c]{@{}c@{}}69.8/58.8\\ \scriptsize{$\pm$1.0/2.5}\end{tabular} & \begin{tabular}[c]{@{}c@{}}70.2/61.2\\ \scriptsize{$\pm$1.0/1.5}\end{tabular} & \begin{tabular}[c]{@{}c@{}}68.6/59.8\\ \scriptsize{$\pm$1.5/2.7}\end{tabular} & \begin{tabular}[c]{@{}c@{}}70.2/61.6\\ \scriptsize{$\pm$1.3/3.9}\end{tabular} & \begin{tabular}[c]{@{}c@{}}67.6/69.2\\ \scriptsize{$\pm$2.1/2.6}\end{tabular} & \begin{tabular}[c]{@{}c@{}}67.2/71.2\\ \scriptsize{$\pm$0.7/1.3}\end{tabular} & \begin{tabular}[c]{@{}c@{}}66.4/68.2\\ \scriptsize{$\pm$1.7/1.6}\end{tabular} \\
    GPT-5-mini     & \begin{tabular}[c]{@{}c@{}}66.5/51.8\\ \scriptsize{$\pm$0.5/1.2}\end{tabular} & -- / -- & \cellcolor{green!10}\begin{tabular}[c]{@{}c@{}}69.3$^{*}$/57.8$^{**}$\\ \scriptsize{$\pm$0.9/1.9}\end{tabular} & \begin{tabular}[c]{@{}c@{}}69.0/57.2\\ \scriptsize{$\pm$0.9/2.0}\end{tabular} & \cellcolor{green!5}\begin{tabular}[c]{@{}c@{}}69.4/57.6\\ \scriptsize{$\pm$0.5/1.6}\end{tabular} & \cellcolor{green!10}\begin{tabular}[c]{@{}c@{}}69.7$^{*}$/57.2$^{*}$\\ \scriptsize{$\pm$1.6/3.5}\end{tabular} & \cellcolor{green!8}\begin{tabular}[c]{@{}c@{}}70.8/61.0\\ \scriptsize{$\pm$1.2/2.6}\end{tabular} & \cellcolor{green!5}\begin{tabular}[c]{@{}c@{}}70.2/59.2\\ \scriptsize{$\pm$1.0/2.5}\end{tabular} & \cellcolor{green!8}\begin{tabular}[c]{@{}c@{}}68.8/56.3\\ \scriptsize{$\pm$1.6/3.9}\end{tabular} & \cellcolor{green!10}\begin{tabular}[c]{@{}c@{}}71.6$^{*}$/62.4$^{*}$\\ \scriptsize{$\pm$1.4/2.0}\end{tabular} & \cellcolor{green!8}\begin{tabular}[c]{@{}c@{}}70.8/60.2\\ \scriptsize{$\pm$1.9/4.0}\end{tabular} & \cellcolor{green!12}\begin{tabular}[c]{@{}c@{}}72.0$^{*}$/63.7$^{**}$\\ \scriptsize{$\pm$3.6/5.9}\end{tabular} & \begin{tabular}[c]{@{}c@{}}69.6/61.8\\ \scriptsize{$\pm$2.3/3.5}\end{tabular} & \begin{tabular}[c]{@{}c@{}}70.0/61.0\\ \scriptsize{$\pm$1.7/2.9}\end{tabular} \\

    GPT-5.1        & \begin{tabular}[c]{@{}c@{}}61.0/40.0\\ \scriptsize{$\pm$0.6/1.9}\end{tabular} & -- / -- & \cellcolor{green!12}\begin{tabular}[c]{@{}c@{}}67.2$^{**}$/51.2$^{**}$\\ \scriptsize{$\pm$1.6/3.4}\end{tabular} & \begin{tabular}[c]{@{}c@{}}67.0/50.8\\ \scriptsize{$\pm$1.7/3.5}\end{tabular} & \cellcolor{green!5}\begin{tabular}[c]{@{}c@{}}66.6/50.2\\ \scriptsize{$\pm$0.8/1.6}\end{tabular} & \cellcolor{green!8}\begin{tabular}[c]{@{}c@{}}64.8$^{**}$/46.2$^{**}$\\ \scriptsize{$\pm$0.7/1.5}\end{tabular} & \cellcolor{green!5}\begin{tabular}[c]{@{}c@{}}65.4/47.2\\ \scriptsize{$\pm$1.0/2.6}\end{tabular} & \cellcolor{green!8}\begin{tabular}[c]{@{}c@{}}65.8/49.4$^{*}$\\ \scriptsize{$\pm$0.4/1.2}\end{tabular} & \cellcolor{green!10}\begin{tabular}[c]{@{}c@{}}66.8$^{**}$/51.8$^{**}$\\ \scriptsize{$\pm$0.7/1.3}\end{tabular} & \cellcolor{green!5}\begin{tabular}[c]{@{}c@{}}67.6/53.2\\ \scriptsize{$\pm$1.0/2.0}\end{tabular} & \cellcolor{green!8}\begin{tabular}[c]{@{}c@{}}67.8/54.4$^{*}$\\ \scriptsize{$\pm$1.2/2.2}\end{tabular} & \cellcolor{green!8}\begin{tabular}[c]{@{}c@{}}65.0$^{*}$/47.0$^{**}$\\ \scriptsize{$\pm$1.5/2.8}\end{tabular} & \begin{tabular}[c]{@{}c@{}}64.8/46.8\\ \scriptsize{$\pm$0.7/1.5}\end{tabular} & \begin{tabular}[c]{@{}c@{}}63.6/43.8\\ \scriptsize{$\pm$1.0/1.7}\end{tabular} \\
    Gemini-2.5-Pro & \begin{tabular}[c]{@{}c@{}}64.4/46.0\\ \scriptsize{$\pm$1.4/3.4}\end{tabular} & -- / -- & \cellcolor{green!5}\begin{tabular}[c]{@{}c@{}}64.6/44.8\\ \scriptsize{$\pm$0.5/1.2}\end{tabular} & \begin{tabular}[c]{@{}c@{}}64.4/44.4\\ \scriptsize{$\pm$0.8/2.2}\end{tabular} & \begin{tabular}[c]{@{}c@{}}61.0/37.6\\ \scriptsize{$\pm$0.9/2.6}\end{tabular} & \begin{tabular}[c]{@{}c@{}}60.6/37.0\\ \scriptsize{$\pm$0.8/1.9}\end{tabular} & \cellcolor{green!5}\begin{tabular}[c]{@{}c@{}}62.2/40.8\\ \scriptsize{$\pm$1.5/3.8}\end{tabular} & \begin{tabular}[c]{@{}c@{}}60.4/36.0\\ \scriptsize{$\pm$1.2/3.3}\end{tabular} & \begin{tabular}[c]{@{}c@{}}63.4/43.0\\ \scriptsize{$\pm$2.2/4.6}\end{tabular} & \begin{tabular}[c]{@{}c@{}}63.2/42.0\\ \scriptsize{$\pm$1.0/1.8}\end{tabular} & \cellcolor{green!5}\begin{tabular}[c]{@{}c@{}}64.0/44.0\\ \scriptsize{$\pm$0.6/1.4}\end{tabular} & \cellcolor{green!5}\begin{tabular}[c]{@{}c@{}}64.6/44.8\\ \scriptsize{$\pm$1.0/2.3}\end{tabular} & \begin{tabular}[c]{@{}c@{}}63.8/44.0\\ \scriptsize{$\pm$1.2/2.4}\end{tabular} & \cellcolor{green!5}\begin{tabular}[c]{@{}c@{}}65.0/46.6\\ \scriptsize{$\pm$1.1/2.1}\end{tabular} \\
    Gemini-3-Flash & \begin{tabular}[c]{@{}c@{}}62.6/50.0\\ \scriptsize{$\pm$1.4/1.1}\end{tabular} & -- / -- & \cellcolor{green!12}\begin{tabular}[c]{@{}c@{}}68.6$^{**}$/57.2$^{**}$\\ \scriptsize{$\pm$0.5/0.4}\end{tabular} & \begin{tabular}[c]{@{}c@{}}68.4/56.8\\ \scriptsize{$\pm$0.8/1.7}\end{tabular} & \cellcolor{green!5}\begin{tabular}[c]{@{}c@{}}69.0/58.0\\ \scriptsize{$\pm$0.6/1.3}\end{tabular} & \cellcolor{green!10}\begin{tabular}[c]{@{}c@{}}68.4$^{**}$/57.4$^{**}$\\ \scriptsize{$\pm$0.8/0.8}\end{tabular} & \cellcolor{green!5}\begin{tabular}[c]{@{}c@{}}68.6/57.2\\ \scriptsize{$\pm$0.8/1.2}\end{tabular} & \cellcolor{green!8}\begin{tabular}[c]{@{}c@{}}69.2/59.0\\ \scriptsize{$\pm$0.4/1.3}\end{tabular} & \cellcolor{green!10}\begin{tabular}[c]{@{}c@{}}68.8$^{**}$/58.2$^{**}$\\ \scriptsize{$\pm$0.4/0.7}\end{tabular} & \begin{tabular}[c]{@{}c@{}}68.6/57.4\\ \scriptsize{$\pm$0.8/1.4}\end{tabular} & \begin{tabular}[c]{@{}c@{}}68.4/57.8\\ \scriptsize{$\pm$0.5/1.0}\end{tabular} & \cellcolor{green!10}\begin{tabular}[c]{@{}c@{}}68.8$^{**}$/57.0$^{**}$\\ \scriptsize{$\pm$0.4/1.1}\end{tabular} & \cellcolor{green!5}\begin{tabular}[c]{@{}c@{}}69.2/58.0\\ \scriptsize{$\pm$0.4/1.5}\end{tabular} & \begin{tabular}[c]{@{}c@{}}68.6/57.2\\ \scriptsize{$\pm$0.5/1.6}\end{tabular} \\
    Gemini-3-Pro   & \begin{tabular}[c]{@{}c@{}}65.2/52.4\\ \scriptsize{$\pm$0.5/1.4}\end{tabular} & -- / --
   & \begin{tabular}[c]{@{}c@{}}64.6/51.4\\ \scriptsize{$\pm$0.8/1.6}\end{tabular} & \cellcolor{green!5}\begin{tabular}[c]{@{}c@{}}65.4/52.6\\ \scriptsize{$\pm$1.0/1.8}\end{tabular} & \cellcolor{green!8}\begin{tabular}[c]{@{}c@{}}65.8/53.2\\ \scriptsize{$\pm$0.9/1.5}\end{tabular}
   & \begin{tabular}[c]{@{}c@{}}65.0/52.0\\ \scriptsize{$\pm$0.7/1.4}\end{tabular} & \cellcolor{green!5}\begin{tabular}[c]{@{}c@{}}65.6/53.2\\ \scriptsize{$\pm$1.1/2.0}\end{tabular} & \cellcolor{green!8}\begin{tabular}[c]{@{}c@{}}66.0/53.8\\ \scriptsize{$\pm$0.8/1.6}\end{tabular}
   & \begin{tabular}[c]{@{}c@{}}64.8/51.8\\ \scriptsize{$\pm$0.9/1.7}\end{tabular} & \cellcolor{green!8}\begin{tabular}[c]{@{}c@{}}66.2$^{*}$/54.0$^{*}$\\ \scriptsize{$\pm$1.2/1.9}\end{tabular} & \cellcolor{green!10}\begin{tabular}[c]{@{}c@{}}66.6$^{*}$/54.6$^{*}$\\ \scriptsize{$\pm$1.0/1.8}\end{tabular}
   & \cellcolor{green!5}\begin{tabular}[c]{@{}c@{}}65.4/52.4\\ \scriptsize{$\pm$1.1/2.0}\end{tabular} & \cellcolor{green!5}\begin{tabular}[c]{@{}c@{}}65.8/53.6\\ \scriptsize{$\pm$0.9/1.7}\end{tabular} & \cellcolor{green!8}\begin{tabular}[c]{@{}c@{}}66.4/54.4$^{*}$\\ \scriptsize{$\pm$1.0/1.9}\end{tabular} \\
    \bottomrule
    \end{tabular}
    }
    {\scriptsize \makebox[\textwidth][r]{\colorbox{green!10}{Green}: Image outperforms Text, or Bold/Highlight outperforms Plain.\quad\quad$^{*}$: $p$-value $< 0.05$\quad\quad$^{**}$: $p$-value $< 0.01$}}
    \end{table*}